\documentclass[journal,10pt]{IEEEtran}
\usepackage{epsfig,latexsym}
\usepackage{float}
\usepackage{indentfirst}
\usepackage{amsmath}
\usepackage{amssymb}
\usepackage{times}
\usepackage{psfrag}
\usepackage{cite}
\usepackage{subfigure}
\usepackage{textcomp}
\usepackage{multirow}
\usepackage{multicol}
\usepackage{stfloats}
\usepackage{url}
\usepackage{subfigure}
\usepackage{multirow}
\usepackage{booktabs}
\usepackage{comment}
\usepackage{threeparttable} % add comment to tables

\begin{document}



\title{Reducing Complexity of HEVC: A Deep Learning Approach}


%\author{Mai~Xu,~\IEEEmembership{Senior Member,~IEEE,}  Tianyi Li~\IEEEmembership{Student Member,~IEEE}, Zulin~Wang, Xin Deng, Ren Yang~\IEEEmembership{Student Member,~IEEE} and Zhenyu Guan
\author{Mai~Xu,~\IEEEmembership{Senior Member,~IEEE,}  Tianyi Li, Zulin Wang, Xin Deng, Ren Yang and Zhenyu Guan
\thanks{M. Xu, T. Li, Z. Wang, R. Yang and Z. Guan are with the School of Electronic and Information Engineering, Beihang University, Beijing, 100191 China. X. Deng is with Department of Department of Electrical and Electronic Engineering, Imperial College London, SW7 2AZ, UK. This work was supported by NSFC under grant number 61573037.}
}
\maketitle

\begin{abstract}
	High Efficiency Video Coding (HEVC) significantly reduces bit-rates over the preceding H.264 standard but at the expense of extremely high encoding complexity.  In HEVC, the quad-tree partition of coding unit (CU) consumes a large proportion of the HEVC encoding complexity, due to the brute-force search for rate-distortion optimization (RDO). Therefore, this paper proposes a deep learning approach to predict the CU partition for reducing the HEVC complexity at both intra- and inter-modes, which is based on convolutional neural network (CNN) and long- and short-term memory (LSTM) network. First, we establish a large-scale database including substantial CU partition data for HEVC intra- and inter-modes. This enables deep learning on the CU partition. Second, we represent the CU partition of an entire coding tree unit (CTU) in the form of a hierarchical CU partition map (HCPM). Then, we propose an early-terminated hierarchical CNN (ETH-CNN) for learning to predict the HCPM. Consequently, the encoding complexity of intra-mode HEVC can be drastically reduced by replacing the brute-force search with ETH-CNN to decide the CU partition. Third, an early-terminated hierarchical LSTM (ETH-LSTM) is proposed to learn the temporal correlation of the CU partition. Then, we combine ETH-LSTM and ETH-CNN to predict the CU partition for reducing the HEVC complexity at inter-mode. Finally, experimental results show that our approach outperforms other state-of-the-art approaches in reducing the HEVC complexity at both intra- and inter-modes.
\end{abstract}


\section{Introduction}
\label{sec:introduction}

The High Efficiency Video Coding (HEVC) standard ~\cite{Sullivan12TCSVT} saves approximately $50\%$ of the bit-rate at similar video quality compared to its predecessor, H.264/Advanced Video Coding (AVC). This is achieved by adopting some advanced video coding techniques, e.g., the quad-tree structure of the coding unit (CU) partition. However, these techniques lead to extremely high complexity of HEVC. As investigated by \cite{Correa12TCSVT}, the encoding time of HEVC is higher than H.264/AVC by $253\%$ on average, making it impractical for implementation in multimedia applications. Therefore, it is necessary to significantly reduce the encoding complexity of HEVC with ignorable loss in rate-distortion (RD) performance.

The past half decade has witnessed a great number of approaches on encoding complexity reduction for HEVC. It is found that the recursive process of the quad-tree CU partition contributes to the largest proportion of the encoding time (at least $80\%$ in the reference software HM \cite{HM}). Thus, most HEVC complexity reduction approaches attempt to simplify the process of the CU partition. The basic idea of these approaches is to predict the CU partition in advance, instead of the brute-force recursive RD optimization (RDO) search.
The early works for predicting the CU partition are heuristic approaches such as in \cite{Leng11CMSP,Xiong14TMM,Shen12PCS}.
These heuristic approaches explore some intermediate features to early determine the CU partition before checking all possible quad-tree patterns.
Since 2015, several machine learning approaches \cite{Correa15TCSVT, Zhang15TIP, Liu16TIP, Mallikarachchi16TCSVT, Zhu17TB} have been proposed to predict the CU partition toward HEVC complexity reduction. For example, to reduce the complexity of inter-mode HEVC, Zhang \textsl{et al.} \cite{Zhang15TIP} proposed a CU depth decision approach with a three-level joint classifier based on support vector machine (SVM), which predicts the splitting of three-sized CUs in the CU partition.
To reduce the complexity of intra-mode HEVC, Liu \textsl{et al.} \cite{Liu16TIP} developed a convolutional neural network (CNN) based approach that predicts the CU partition. However, the structure of the CNN in \cite{Liu16TIP} is shallow, with limited learning capacity, and thus, it is insufficient to accurately model the complicated CU partition process. In this paper, we propose a deep CNN structure for learning to predict the intra-mode CU partition, thus reducing the complexity of HEVC at intra-mode.
In addition, all existing approaches, including \cite{Zhang15TIP} and \cite{Liu16TIP}, do not explore the correlation of the CU partition across neighboring frames. This paper develops a long- and short-term memory (LSTM) structure to learn the temporal dependency of the inter-mode CU partition. Then, a deep learning approach is proposed to predict the CU partition at inter-mode, which combines the CNN and LSTM structures. Consequently, the encoding complexity of inter-mode HEVC can be significantly reduced.

% QP=quantization parameter 	
Specifically, in this paper, we first establish a large-scale database for the CU partition, boosting studies on deep learning based complexity reduction in HEVC. Our database is available online: \url{https://github.com/HEVC-Projects/CPH}. Our database contains the data of the CU partition at both intra-mode (2000 raw images compressed at four QPs) and inter-mode (111 raw sequences compressed at four QPs). Next, we propose a deep learning based complexity reduction approach for both intra- and inter-mode encoding of HEVC, which learns from our database to split an entire coding tree unit (CTU). More specifically, we propose to efficiently represent the CU partition of a CTU through the hierarchical CU partition map (HCPM). Given sufficient training data and efficient HCPM representation, the deep learning structures of our approach are able to ``go deeper'', such that extensive parameters can be learned for exploring diverse patterns of the CU partition.
Accordingly, our deep learning approach introduces an early-terminated hierarchical CNN (ETH-CNN) structure, which hierarchically generates the structured HCPM with an early-termination mechanism.
The early-termination mechanism introduced in ETH-CNN can save computational time when running the CNN.
ETH-CNN can be used to reduce the encoding complexity of intra-mode HEVC. Our approach further introduces an early-terminated hierarchical LSTM (ETH-LSTM) structure for inter-mode HEVC. In ETH-LSTM, the temporal correlation of the CU partition is learned in LSTM cells. With the features of the ETH-CNN as the input, ETH-LSTM hierarchically outputs HCPM by incorporating the learnt LSTM cells with the early-termination mechanism. As such, our approach can also be used to reduce the encoding complexity of inter-mode HEVC.

% add it in enough sapce!!! However, the complexity reduction is at the cost of RD performance loss, due to the inaccuracy of HCPM prediction.  Therefore, a bi-threshold decision scheme is developed to achieve the trade-off between encoding complexity and RD performance.

This paper was previously presented at the conference \cite{Li17ICME}, with the following extensions. Rather than the three-level CU splitting labels of \cite{Li17ICME}, HCPM is proposed in this paper to hierarchically represent the structured output of the CU partition. Based on the proposed HCPM representation, we advance the deep CNN structure of \cite{Li17ICME} by introducing the early-termination mechanism, for intra-mode HEVC complexity reduction. More importantly, this paper further proposes ETH-LSTM to reduce the HEVC complexity at inter-mode.
In contrast, our previous work \cite{Li17ICME} only addresses complexity reduction in intra-mode HEVC.
For learning ETH-LSTM, a large-scale database of inter-mode CU partition is established in this paper, whereas \cite{Li17ICME} only contains the database of intra-mode CU partition.
In brief, the main contributions of this paper are summarized as follows.

\begin{itemize}
	\item
	We establish a large-scale database for the CU partition of HEVC at both intra- and inter-modes, which may facilitate the applications of deep learning in reducing HEVC complexity.
	\item
	We propose a deep CNN structure called ETH-CNN to predict the structured output of the CU partition in the form of HCPM, for reducing the complexity of intra-mode HEVC.
	\item
	We develop a deep LSTM structure named ETH-LSTM that learns the spatio-temporal correlation of the CU partition, for reducing the complexity of HEVC at inter-mode.
\end{itemize}

This paper is organized as follows.
Section II reviews the related works on HEVC complexity reduction.
Section III presents the established CU partition database.
In Sections IV and V, we propose ETH-CNN and ETH-LSTM to reduce the HEVC complexity at intra-mode and inter-mode, respectively.
Section VI reports the experimental results, and Section VII concludes this paper.

\section{Related Works}
\label{sec:related-works}

The existing HEVC complexity reduction works can be generally classified into two categories: heuristic and learning-based approaches. This section reviews the complexity reduction approaches in these two categories.

In heuristic approaches, the brute-force RDO search of the CU partition can be simplified according to some intermediate features. The representative approaches include \cite{Leng11CMSP,Xiong14TMM,Cho13TCSVT,Shen12PCS,Kim16BMSB}.
To be more specific, Leng \textsl{et al.}~\cite{Leng11CMSP} proposed a CU depth decision approach at  the frame level, which skips certain CU depths rarely found in the previous frames.
At the CU level, Xiong \textsl{et al.}~\cite{Xiong14TMM} and Kim \textsl{et al.}~\cite{Kim16BMSB} proposed deciding whether to split CUs based on the pyramid motion divergence and the number of high-frequency key-points, respectively.
Shen \textsl{et al.}~\cite{Shen12PCS} proposed selecting some crucial and computational-friendly features, e.g., RD cost and inter-mode prediction error, to early determine the splitting of each CU. Then, the CU splitting is determined based on the rule of minimizing the Bayesian risk.
Also based on the Bayesian rule, Cho \textsl{et al.}~\cite{Cho13TCSVT} developed a CU splitting and pruning approach according to the features of full and low-complexity RD costs.
In addition to simplified the CU partition, various heuristic approaches \cite{Khan13ICIP, Yoo13ICCE, Kiho12EL, Cui17TIP} were proposed to reduce the complexity of prediction unit (PU) or transform unit (TU) partition.
For example, Khan \textsl{et al.}~\cite{Khan13ICIP} proposed a fast PU size decision approach, which adaptively integrates smaller PUs into larger PUs with regard to video content.
Yoo \textsl{et al.} \cite{Yoo13ICCE} proposed estimating the PU partition with the maximum probability, on account of the coding block flag (CBF) and RD costs of encoded PUs.
In the latest work of \cite{Cui17TIP}, the RDO quantization (RDOQ) process is accelerated based on the transform coefficients with a hybrid Laplace distribution.  In addition, other components of HEVC, such as intra- and inter-mode prediction and in-loop filtering, are simplified to reduce the encoding complexity in \cite{Miyazawa12PCS, Zhang14TCSVT, Vanne14TCSVT, Shen14TIP, Park15TIP}.

Most recently, learning-based approaches \cite{Correa15TCSVT, Hu16BMSB, Liu16DASC, Zhang15TIP, Zhu17TB, Liu16TIP, Laude16PCS_Deep, Kim16TCSVT, Tohidypour16TCSVT, Oliveira16EL, Duanmu16JESTCS, Momcilovic15ISM, Du15APSIPA, Shan17ICASSP, Mallikarachchi16TCSVT} have flourished for complexity reduction in HEVC. These approaches utilize  machine learning from extensive data to generalize the rules of HEVC encoding components, instead of a brute-force RDO search on these components.
For example, for intra-mode HEVC, \cite{Hu16BMSB} modeled the CU partition process as a binary classification problem with logistic regression, and \cite{Liu16DASC} proposed using SVM to perform classification in the CU partition process.
As a result, the computational time of the CU partition can be decreased using well-trained classification models instead of brute-force search.
For inter-mode HEVC, Corr\^ea \textsl{et al.}~\cite{Correa15TCSVT} proposed three early termination schemes with data mining techniques to simplify the decision on the optimal CTU structures.
In \cite{Zhang15TIP}, several HEVC domain features that are correlated with the CU partition were explored. Then, a joint classifier of SVM was proposed to utilize these features in determining CU depths, such that the encoding complexity of HEVC can be reduced because the brute-force RDO search is skipped. Later, a binary and multi-class SVM algorithm \cite{Zhu17TB} was proposed to predict both the CU partition and PU mode with an off-on-line machine learning mechanism. % that is from \cite{Zhu17TB}
As such, the encoding time of HEVC can be further saved.
However, the above learning-based approaches rely heavily on hand-crafted features that are related to the CU partition.

For complexity reduction in HEVC, the CU partition-related features can be automatically extracted by deep learning, rather than manual extraction. The advantage of applying deep learning for HEVC complexity reduction is that it can take advantage of large-scale data to automatically mine extensive features related to the CU partition, rather than the limited hand-crafted features. Unfortunately, there are few deep learning works on the prediction of the CU partition.
To the best of our knowledge, the only approaches in this direction are \cite{Liu16TIP, Laude16PCS_Deep}, in which the CNN architectures have been developed to predict intra-mode CU partition. The CNN architectures of \cite{Liu16TIP, Laude16PCS_Deep} are not sufficiently deep due to a lack of training data. For example,  \cite{Liu16TIP} only contains two convolutional layers with 6 and 16 kernels sized $3\times3$.
%To avoid such a disadvantage, this paper establishes a large-scale database of the CU partition. Our database enables a deeper CNN architecture, ETH-CNN, such that the prediction accuracy of the CU partition can be significantly improved. Meanwhile, the computational time of ETH-CNN can be reduced by introducing  the non-overlapping convolution and early termination mechanism. In addition, ETH-LSTM is proposed in this paper to consider the temporal correlation in predicting the CU partition to reduce the HEVC complexity at inter-mode, whereas both \cite{Liu16TIP} and \cite{Laude16PCS_Deep} only handle HEVC complexity reduction at intra-mode.
In contrast, we propose the deep architectures of ETH-CNN and ETH-LSTM for HEVC complexity reduction at both intra- and inter-modes.

Our approach differs from the above learning-based CU partition prediction approaches in four aspects.
(1) Compared with the conventional three-level CU splitting labels as in [8], [9], [11], [26], the HCPM is proposed to efficiently represent the structured output of the CU partition. This can dramatically reduce the computational time of the CU partition, since the complete CU partition in a whole CTU can be obtained in terms of one HCPM by running the trained ETH-CNN/ETH-LSTM model only once.
(2) A deep ETH-CNN structure is designed to automatically extract features for predicting the CU partition, instead of the handcrafted feature extraction in [8], [11].
Besides, the deep ETH-CNN structure has much more trainable parameters than the CNN structures of [9] and [26], thus remarkably improving the prediction accuracy of the CU partition. Additionally, the early-termination mechanism proposed in our approach can further save computational time.
(3) A deep ETH-LSTM model is further developed for learning long- and short-term dependencies of the CU partition across frames for inter-mode HEVC. To our best knowledge, this is a first attempt to leverage LSTM for predicting CU partition in HEVC complexity reduction.
(4) To train the large amount of parameters in ETH-CNN and ETH-LSTM, we establish a large-scale database for the CU partition of HEVC at both intra- and inter-modes. In contrast, other works only rely on the existing JCT-VC database, much smaller than our database. Our database may facilitate the future works of applying deep learning in the CU partition prediction for reducing HEVC complexity.

\section{CU Partition Database}
\label{sec:ctu_database}

\subsection{Overview of CU Partition}
\label{sec:ctu}
The CTU partition structure ~\cite{Sullivan12TCSVT} is one of the major contributions to the HEVC standard, with the CU partition as the core process.
The size of a CTU is 64$\times$64 pixels by default, and a CTU can either contain a single CU or be recursively split into multiple smaller CUs, based on the quad-tree structure.
The minimum size of a CU is configured before encoding, with the default being 8$\times$8.
Thus, the sizes of the CUs in the CTU are diverse, ranging from 64$\times$64 to 8$\times$8. Note that the maximal CTU sizes can be extended to be larger than 64$\times$64, e.g., 128$\times$128.

\begin{figure}
	\centering
	\includegraphics[width=75mm]{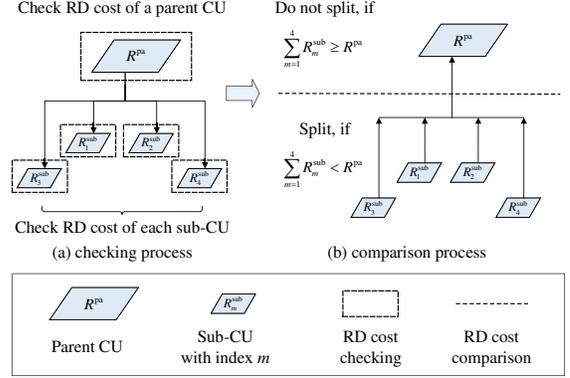}\\
    \caption{\footnotesize{Illustration of checking and comparing RD cost between a parent CU and its sub-CUs. Note that this illustration can be applied to the splitting of $64 \times 64 \rightarrow 32 \times 32$, $32 \times 32 \rightarrow 16 \times 16$ or $16 \times 16 \rightarrow 8 \times 8$.}}
	\label{fig:ctu}
\end{figure}

As we know, the sizes of the CUs in each CTU are determined using a brute-force RDO search, which includes a top-down checking process and a bottom-up comparison process.
Fig. \ref{fig:ctu} illustrates the RD cost checking and comparison between a parent CU and its sub-CUs.
In the checking process, the encoder checks the RD cost of the whole CTU, followed by checking its sub-CUs, until reaching the minimum CU size.
In Fig. \ref{fig:ctu}, the RD cost of a parent CU is denoted as $R^{\text{pa}}$, and the RD costs of its sub-CUs are denoted as $R^{\text{sub}}_{m}$, where $m\in\{1,2,3,4\}$ is the index of each sub-CU.
Afterwards, based on the RD costs of CUs and sub-CUs, the comparison process is conducted to determine whether a parent CU should be split.
As shown in Fig. \ref{fig:ctu} (b), if $\sum_{m=1}^{4}R^{\text{sub}}_{m}<R^{\text{pa}}$, the parent CU needs to be split; otherwise, it is not split. Note that the RD cost of the split flag is also considered when deciding whether to split the CU.
After the full RDO search, the CU partition with the minimum RD cost is selected.

It is worth noting that the RDO search is extremely time-consuming, mainly attributed to the recursive checking process.
In a 64$\times$64 CTU, 85 possible CUs are checked, including $1$, $4$, $4^2$ and $4^3$ CUs with sizes of 64$\times$64, 32$\times$32, 16$\times$16 and 8$\times$8.
To check the RD cost of each CU, the encoder needs to execute pre-coding for the CU, in which the possible prediction and transformation modes have to be encoded.
More importantly, the pre-coding has to be conducted for all 85 possible CUs in the standard HEVC, consuming the largest proportion of the encoding time.
However, in the final CU partition, only certain CUs are selected, from 1 (if the 64$\times$64 CU is not split) to 64 (if the whole CTU is split into 8$\times$8 CUs).
Therefore, the pre-coding of 84 CUs (i.e., 85-1) at most and 21 CUs (i.e., 85-64) at least can be avoided through the accurate prediction of the CU partition.

\subsection{Database Establishment for Intra-mode}
\label{sec:database-intra}

In this section, we present our large-scale database for the CU Partition of HEVC at Intra-mode: the CPH-Intra database. To the best of our knowledge, our database is the first database on the CU partition patterns. To establish the CPH-Intra database, 2000 images at resolution 4928$\times$3264 were selected from the Raw Images Dataset (RAISE) \cite{Dang2015MM_RAISE}. These 2000 images were randomly divided into training (1700 images), validation (100 images) and test (200 images) sets. Furthermore, each set was equally divided into four subsets: one subset was with original resolution, and the other three sets were down-sampled to 2880$\times$1920, 1536$\times$1024 and 768$\times$512. As such, our CPH-Intra database contains images at different resolutions. This ensures sufficient and diverse training data for learning to predict the CU partition.

Next, all images were encoded by the HEVC reference software HM 16.5 \cite{HM}. Here, four QPs $\{22,27,32,37\}$ were applied to encode at the All-Intra (AI) configuration with the file \emph{encoder\_intra\_main.cfg} \cite{TestCond13}.
After encoding, the binary labels indicating splitting (=1) and non-splitting (=0) can be obtained for all CUs, and each CU with its corresponding binary label is a sample in our database.
Finally, there are 110,405,784 samples collected in the CPH-Intra database. These samples are split into 12 sub-databases according to their QP value and CU size, as reported in Table I-(a) of the \textit{Supporting Document}. As shown in this table, the total percentage of split CUs (49.2\%) is close to that of non-split CUs (50.8\%) over the whole database.

\subsection{Database Establishment for Inter-mode}
\label{sec:database-inter}
We further establish a database for the CU Partition of HEVC at Inter-mode: the CPH-Inter database. For establishing the CPH-Inter database, 111 raw video sequences were selected, therein consisting of 6 sequences at 1080p (1920$\times$1080) from \cite{Xu2014JSTP}, 18 sequences of Classes A $\sim$ E from the Joint Collaborative Team on Video Coding (JCT-VC) standard test set\cite{Ohm12TCSVT}, and 87 sequences from Xiph.org \cite{XIPH2017} at different resolutions. As a result, our CPH-Inter database contains video sequences at various resolutions: SIF (352$\times$240), CIF (352$\times$288), NTSC (720$\times$486), 4CIF (704$\times$576), 240p (416$\times$ 240), 480p (832$\times$480), 720p (1280$\times$720), 1080p, WQXGA (2560$\times$1600) and 4K (4096$\times$2160). Note that the NTSC sequences were cropped to 720$\times$480 by removing  the bottom edges of the frames, considering that only resolutions in multiples of 8$\times$8 are supported. Moreover, if the durations of the sequences are longer than 10 seconds, they were clipped to be 10 seconds.

In our CPH-Inter database, all the above sequences were divided into non-overlapping training (83 sequences), validation (10 sequences) and test (18 sequences) sets. For the test set, all 18 sequences from the JCT-VC set \cite{Ohm12TCSVT} were selected.
Similar to the CPH-Intra database, all sequences in our CPH-Inter database were encoded by HM 16.5 \cite{HM} at four QPs $\{22,27,32,37\}$.
They were compressed with the Low Delay P (LDP) (using \textit{encoder\_lowdelay\_P\_main.cfg}), the Low Delay B (LDB) (using \textit{encoder\_lowdelay\_main.cfg}) and the Random Access (RA) (using \textit{encoder\_randomaccess\_main.cfg}) configurations, respectively.
Consequently, 12 sub-databases were obtained for each configuration, corresponding to different QPs and CU sizes. As reported in Table I-(b), -(c) and -(d) of the \textit{Supporting Document}, totally 307,831,288 samples were collected for the LDP configuration in our CPH-Inter database. Besides, 275,163,224 and 232,095,164 samples were collected in total in our database for the LDB and RA configurations, respectively.

\section{Complexity Reduction for Intra-mode HEVC}
\label{sec:method-intra}

\subsection{Hierarchical CU Partition Map}
\label{sec:classifier}

\begin{figure}	
	\centering
	\includegraphics[width=85mm]{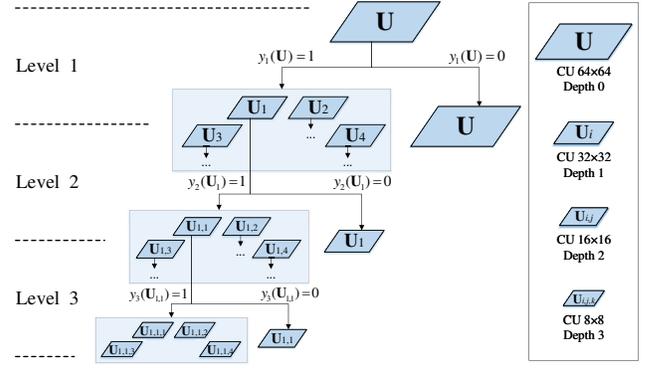}\\
	\caption{\footnotesize{Illustration of three-level CU classifier.}
	}
	\label{fig:classifier}
\end{figure}

According to the CU partition structure in HEVC, four different CU sizes are supported by default, i.e., $64\times64$, $32\times32$, $16\times16$ and $8\times8$, corresponding to four CU depths, 0, 1, 2 and 3. Note that a CU of size $\geq16\times16$ can be either split or not split. As illustrated in Fig. \ref{fig:classifier}, the overall CU partition can be regarded as a combination of binary labels $\{y_l\}_{l=1}^3$ at three levels, where $l\in\{1,2,3\}$ represents the level on which the split decision is made. In particular, $l=1$ indicates the first level, which determines whether to split a $64\times64$ CU into $32\times32$ CUs. Similarly, $l=2$ is the second level to decide whether a $32\times32$ CU is split into $16\times16$ CUs, and $l=3$ is for  $16\times16$ into $8\times8$.

Given a CTU, we assume that the CU of depth 0 is denoted as $\mathbf{U}$.
For $\mathbf{U}$, the first-level label $y_1(\mathbf{U})$ indicates whether $\mathbf{U}$ is split ($=1$) or not ($=0$).
If $\mathbf{U}$ is split, its sub-CUs of depth 1 are denoted as $\{\mathbf{U}_i\}_{i=1}^4$.
Then, the second-level labels $\{y_2(\mathbf{U}_i)\}_{i=1}^4$ denote whether $\{\mathbf{U}_i\}_{i=1}^4$ are split ($=1$) or not ($=0$).
For each split $\mathbf{U}_i$, its sub-CUs of depth 2 are represented by $\{\mathbf{U}_{i,j}\}_{j=1}^4$.
Similarly, the labels $\{y_3(\mathbf{U}_{i,j})\}_{i,j=1}^4$ denote the split labels of $\{\mathbf{U}_{i,j}\}_{i,j=1}^4$ at the third level, and for each split $\mathbf{U}_{i,j}$, its sub-CUs of depth 3 are $\{\mathbf{U}_{i,j,k}\}_{k=1}^4$.
The subscripts $i,j,k\in\{1,2,3,4\}$ are the indices of the sub-CUs split from $\mathbf{U}$, $\mathbf{U}_i$ and $\mathbf{U}_{i,j}$, respectively.
The above hierarchical labels for the CU split are represented by the downward arrows in Fig. \ref{fig:classifier}. The overall CU partition in a CTU is extremely complex, due to the large number of possible pattern combinations.
For example, for a $64\times64$ $\mathbf{U}$, if $y_1(\mathbf{U})=1$, it will be split into four $32\times32$ CUs, i.e., $\{\mathbf{U}_i\}_{i=1}^4$. Since for each $\mathbf{U}_i$ there exist $1+2^4=17$ splitting patterns in $\{\mathbf{U}_{i,j}\}_{j=1}^4$, the total number of splitting patterns for $\mathbf{U}$ is $1+17^4=83522$.

In HEVC, the labels $\{y_l\}_{l=1}^3$ are obtained with the time-consuming RDO process, as mentioned in Section \ref{sec:ctu}. They can be predicted at a much faster speed by machine learning. However, due to the enormous patterns of the CU partition (83,522 patterns, as discussed above), it is intractable to predict the CU partition patterns by a simple multi-class classification in one step.
Instead, the prediction should be made step by step in a hierarchy to yield $\hat{y}_1(\mathbf{U})$, $\{\hat{y}_2(\mathbf{U}_i)\}_{i=1}^4$ and $\{\hat{y}_3(\mathbf{U}_{i,j})\}_{i,j=1}^4$, which denote the predicted $y_1(\mathbf{U})$, $\{y_2(\mathbf{U}_i)\}_{i=1}^4$ and $\{y_3(\mathbf{U}_{i,j})\}_{i,j=1}^4$, respectively.

In typical machine learning based methods \cite{Hu16BMSB,Liu16DASC,Zhang15TIP,Liu16TIP}, the binary labels $\hat{y}_1(\mathbf{U})$, $\{\hat{y}_2(\mathbf{U}_i)\}_{i=1}^4$ and  $\{\hat{y}_3(\mathbf{U}_{i,j})\}_{i,j=1}^4$ are predicted separately for CUs of size $64\times64$, $32\times32$ and $16\times16$. To determine the CU partition for an entire CTU, the trained model needs to be invoked multiple times, leading to repetitive computation overhead.
To overcome such a drawback, we develop an HCPM to efficiently represent CU splitting as the structured output of machine learning, such that the training model is run only once for predicting the partition of the whole CTU.
This way, the computational time of determining the CU partition can be dramatically reduced. In addition, the encoding complexity can be decreased remarkably by bypassing the redundant RD cost checking in the conventional HEVC standard.

\begin{figure}
	\centering
	\includegraphics[width=85mm]{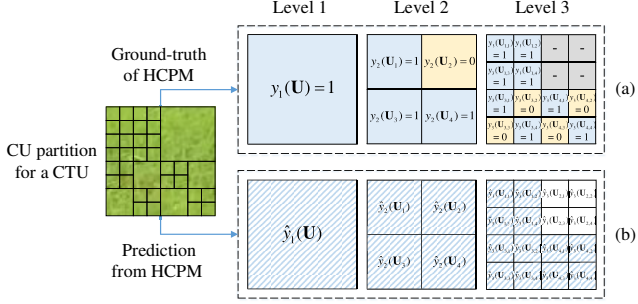}\\
	\caption{\footnotesize{
			An example of HCPM. (a) The ground-truth of HCPM for CU splitting. (b) Modeling HCPM for prediction the CU splitting.
	}}
	\label{fig:truth-and-pred}
\end{figure}

Fig. \ref{fig:truth-and-pred} shows an example of HCPM, which arranges the labels of the CU splitting as the structured output in a hierarchy.
Specifically, HCPM hierarchically contains $1\times 1$, $2\times 2$ and $4\times 4$ binary labels at levels 1, 2 and 3, respectively, corresponding to $y_1(\mathbf{U})$, $\{y_2(\mathbf{U}_i)\}_{i=1}^4$ and $\{y_3(\mathbf{U}_{i,j})\}_{i,j=1}^4$ for ground-truth and $\hat{y}_1(\mathbf{U})$, $\{\hat{y}_2(\mathbf{U}_i)\}_{i=1}^4$ and $\{\hat{y}_3(\mathbf{U}_{i,j})\}_{i,j=1}^4$ for prediction, respectively. Note that when $\mathbf{U}$ or $\mathbf{U}_i$ is  not split, the corresponding sub-CUs $\{\mathbf{U}_i\}_{i=1}^4$ or $\{\mathbf{U}_{i,j}\}_{j=1}^4$ do not exist. Accordingly, the labels of $\{y_2(\mathbf{U}_i)\}_{i=1}^4$ or $\{y_3(\mathbf{U}_{i,j})\}_{j=1}^4$ are set to be null, denoted as ``-'' in the HCPM.

\vspace{-0.5em}
\subsection{ETH-CNN Structure for Learning HCPM}
\label{sec:cnn_struct}

\begin{figure*}
	\centering
	\includegraphics[width=135mm]{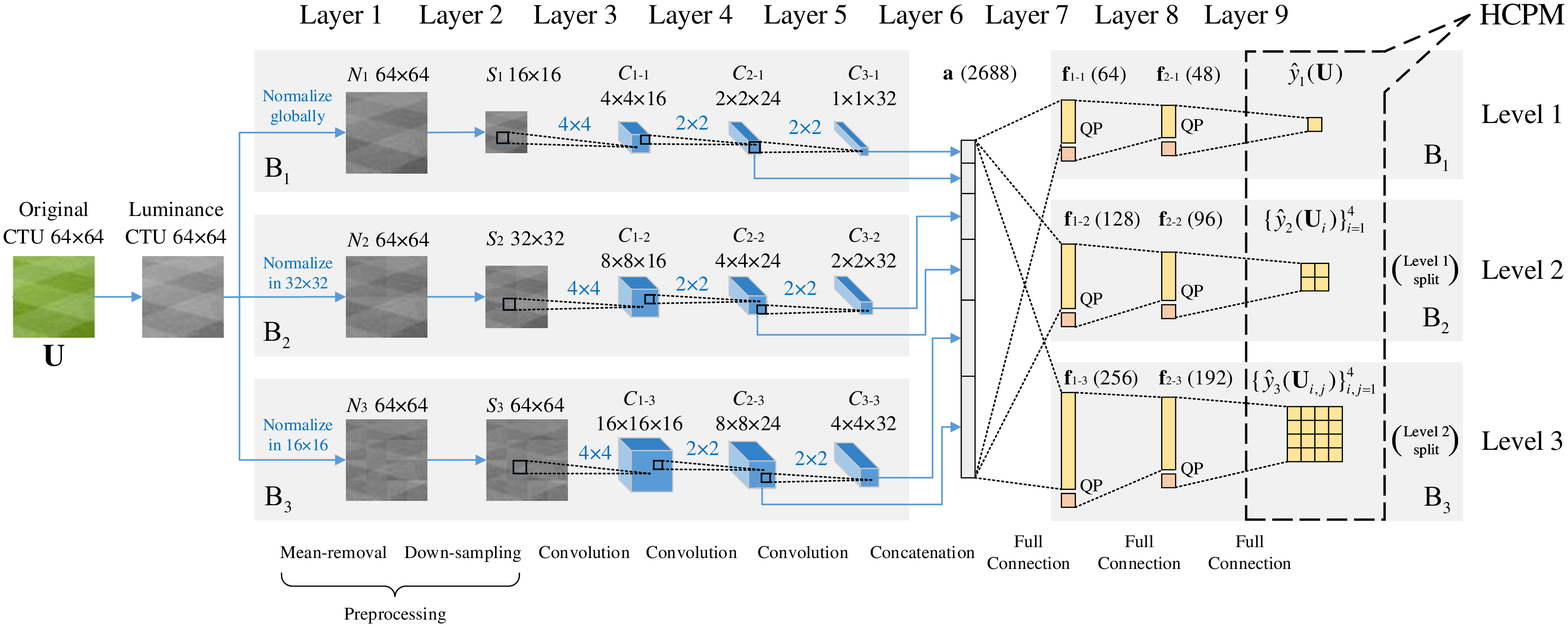}\\
	\caption{\footnotesize{
			Structure of ETH-CNN.
			For preprocessing and convolution, the size of the output at each layer is shown in black font,
			%e.g., $S_1$ is with size $16\times 16$, and the size of $C_{1-1}$ is $4\times 4 \times 16$. Besides,
			and the blue font indicates the kernel size for convolution.
			%, e.g., kernel size for generating $C_{1-1}$ is $4 \times 4$.
			At concatenation and fully connected layers, the number inside each bracket represents the vector length.
		}
	}
	\label{fig:cnn}
\end{figure*}

Next, a deep ETH-CNN structure is developed to learn HCPM for predicting the CU partition of intra-mode HEVC. According to the mechanism of the CU partition, the ETH-CNN structure is designed in Fig. \ref{fig:cnn}. We can see from this figure that ETH-CNN is fed with an entire CTU, denoted by $\mathbf{U}$, and is able to hierarchically generate three branches of structured output, representing all predictions
$\hat{y}_1(\mathbf{U})$, $\{\hat{y}_2(\mathbf{U}_i)\}_{i=1}^4$ and $\{\hat{y}_3(\mathbf{U}_{i,j})\}_{i,j=1}^4$ of HCPM at three levels.
Note that the input CTU is extracted from raw images or sequences in YUV format, and only the Y channel is used in ETH-CNN as this channel contains most visual information.
In contrast to the conventional CNN structure, the mechanism of early termination is introduced in ETH-CNN, which may early terminate the three fully connected layers at the second and third branches.
In addition to the three fully connected layers, the ETH-CNN structure is composed of two preprocessing layers, three convolutional layers, and one concatenating layer.
In the following, we briefly present the details of these layers.

\begin{table}[t]
	\scriptsize
	\newcommand{\tabincell}[2]{\begin{tabular}{@{}#1@{}}#2\end{tabular}}
	\begin{center}
		\caption{\footnotesize{Configuration of proposed ETH-CNN}} \label{tab:cnn}
		\begin{tabular}{|c|c|c|c|c|c|c|}
			%\hline \tabincell{c}{Feature Map\\/ Vector} & Size & \tabincell{c}{Number of\\Trainable Parameters} & \tabincell{c}{Number of\\Additions} & \tabincell{c}{Number of\\Multiplications}\\
			\hline \tabincell{c}{Feature} & Size & \tabincell{c}{Number of\\Parameters} & \tabincell{c}{Number of\\Additions} & \tabincell{c}{Number of\\Multiplications}\\
			\hline $C_{1-1}$ & 16$\times{}$16 & 256 & 3,840 & 4,096\\
			\hline $C_{1-2}$ & 32$\times{}$32 & 256 & 15,360 & 16,384\\
			\hline $C_{1-3}$ & 64$\times{}$64 & 256 & 61,440 & 65,536\\
			\hline $C_{2-1}$ & 4$\times{}$4 & 1,536 & 4,608 & 6,144\\
			\hline $C_{2-2}$ & 8$\times{}$8 & 1,536 & 18,432 & 24,576\\
			\hline $C_{2-3}$ & 16$\times{}$16 & 1,536 & 73,728 & 98,304\\
			\hline $C_{3-1}$ & 2$\times{}$2 & 3,072 & 2,304 & 3,072\\
			\hline $C_{3-2}$ & 4$\times{}$4 & 3,072 & 9,216 & 12,288\\
			\hline $C_{3-3}$ & 8$\times{}$8 & 3,072 & 36,864 & 49,152\\
			\hline $\mathbf{f}_{1-1}$ & 2688 & 172,032 & 171,968 & 172,032\\
			\hline $\mathbf{f}_{1-2}$ & 2688 & 344,064 & 343,936 & 344,064\\
			\hline $\mathbf{f}_{1-3}$ & 2688 & 688,128 & 687,872 & 688,128\\
			\hline $\mathbf{f}_{2-1}$ & 65 & 3,120 & 3,072 & 3,120\\
			\hline $\mathbf{f}_{2-2}$ & 129 & 12,384 & 12,288 & 12,384\\
			\hline $\mathbf{f}_{2-3}$ & 257 & 49,344 & 49,152 & 49,344\\
			\hline $\hat{y}_1(\mathbf{U})$ & 49 & 49 & 48 & 49\\
			\hline $\{\hat{y}_2(\mathbf{U}_i)\}_{i=1}^4$ & 97 & 388 & 384 & 388\\
			\hline $\{\hat{y}_3(\mathbf{U}_{i,j})\}_{i,j=1}^4$ & 193 & 3,088 & 3,072 & 3,088\\
			\hline Total & - & 1,287,189 & 1,497,584 & 1,552,149 \\
			\hline
		\end{tabular}
	\end{center}
\end{table}

\begin{itemize}
	\item
	\textbf{Preprocessing layers.}
	The raw CTU is preprocessed by mean removal and down-sampling in three parallel branches $\{\textrm{B}_{l}\}_{l=1}^3$, corresponding to three levels of HCPM.
	Fig. \ref{fig:cnn} denotes mean removal and down-sampling by $\{{N_l}\}_{l=1}^{3}$ and $\{S_l\}_{l=1}^{3}$, respectively.
	For mean removal, at each branch, the input CTUs are subtracted by the mean intensity values to reduce the variation of the input CTU samples. Specifically, at branch $\textrm{B}_{1}$, the mean value of $\mathbf{U}$ is removed in accordance with the single output of $\hat{y}_1(\mathbf{U})$ at the first level of HCPM.
	At branch $\textrm{B}_{2}$, four CUs $\{\mathbf{U}_i\}_{i=1}^4$ are subtracted by their corresponding mean values, matching the $2 \times 2$ output of $\{\hat{y}_2(\mathbf{U}_i)\}_{i=1}^4$ at the second level of HCPM.
	Similarly, $\{\mathbf{U}_{i,j}\}_{i,j=1}^4$ at $\textrm{B}_{3}$ also remove the mean values in each CU for the $4 \times 4$ output $\{\hat{y}_3(\mathbf{U}_{i,j})\}_{i,j=1}^4$ at the third level of HCPM.
	Next, because CTUs of smaller depths generally possess a smooth texture, $\mathbf{U}$ and $\{\mathbf{U}_i\}_{i=1}^4$ at branches $\textrm{B}_{1}$ and $\textrm{B}_{2}$ are down-sampled to $16\times16$ and $32\times32$, respectively.
	This way, the outputs of the three-layer convolutions at branches $\textrm{B}_{1}$ and $\textrm{B}_{2}$ are with the same sizes of HCPM at levels 1 and 2, i.e., $1 \times 1$ at level 1 and $2\times 2$ at level 2.
	
	\item
	\textbf{Convolutional layers.}
	In each branch $\textrm{B}_{l}$, all preprocessed data flow through three convolutional layers. At each convolutional layer, the kernel sizes are the same across different branches. Specifically, the data are convoluted with $4\times4$ kernels (16 filters in total) at the first convolutional layer to extract the low-level features $\{C_{1-l}\}_{l=1}^{3}$ for the CU partition. Recall that $l$ denotes the level of the CU partition. At the second and third layers, feature maps are sequentially convoluted twice with $2\times2$ kernels (24 filters for the second layer and 32 filters for the third layer) to generate features at a higher level, denoted by $\{C_{2-l}\}_{l=1}^{3}$ and $\{C_{3-l}\}_{l=1}^{3}$, respectively.
	We set the strides of all the above convolutions equal to the widths of the corresponding kernels for non-overlap convolutions, in accordance with non-overlap CU splitting.
	
	\item
	\textbf{Concatenating layer.}
	All feature maps at three branches, yielded from the second and third convolutional layers, are concatenated together and then flattened into a vector $\mathbf{a}$. As shown in Fig. \ref{fig:cnn}, the vectorized features of the concatenating layer are gathered from 6 sources of feature maps in total, i.e., $\{C_{2-l}\}_{l=1}^3$ and $\{C_{3-l}\}_{l=1}^3$, to obtain a variety of both global and local features.  With the whole concatenating vector, features generated from the whole CTU are all considered in the following fully connected layers, rather than those only from the CUs in one branch, to predict the CU partition of HCPM at each single level.
	
	\item
	\textbf{Fully connected layers.}
	Next, all features in the concatenated vector are processed in three branches $\{\textrm{B}_l\}_{l=1}^3$, also corresponding to three levels of HCPM. In each branch, the vectorized features of the concatenating layer pass through three fully connected layers, including two hidden layers and one output layer.
	The two hidden fully connected layers successively generate feature vectors $\{\mathbf{f}_{1-l}\}_{l=1}^{3}$, and the output layer produces HCPM as the output of ETH-CNN.
	The numbers of features at these layers vary with respect to $l$, so that the outputs have 1, 4 and 16 elements in $\textrm{B}_1$, $\textrm{B}_2$ and $\textrm{B}_3$, serving as the predicted binary labels of HCPM at the three levels shown in Fig. \ref{fig:truth-and-pred} (b), i.e., $\hat{y}_1(\mathbf{U})$  in $1 \times 1$, $\{\hat{y}_2(\mathbf{U}_i)\}_{i=1}^4$ in $2 \times 2$ and $\{\hat{y}_3(\mathbf{U}_{i,j})\}_{i,j=1}^4$ in $4 \times 4$. Moreover, QP also has a significant influence on the CU partition. As QP decreases, more CUs tend to be split, and vice versa. Therefore, QP is supplemented as an external feature in the feature vectors $\{\mathbf{f}_{1-l}\}_{l=1}^{3}$ for the full connection, enabling ETH-CNN to be adaptive to various QPs in predicting HCPM.
	In ETH-CNN, the early termination may result in the calculation of the fully connected layers at levels 2 and 3 being skipped, thus saving computation time.
	Specifically, if $\mathbf{U}$ is decided not to be split at level 1, the calculation of $\{\hat{y}_2(\mathbf{U}_i)\}_{i=1}^4$ of HCPM is terminated early at level 2.
	If $\{\mathbf{U}_i\}_{i=1}^4$ are all  not split, the $\{\hat{y}_3(\mathbf{U}_{i,j},t)\}_{i,j=1}^4$ at level 3 do not need to be calculated for the early termination.
	%Therefore, early termination is able to save the computational time of ETH-CNN.
	
	\item
	\textbf{Other layers.}
	During the training phase, after the first and second fully connected layers, features are randomly dropped out \cite{Hinton2011CS_Dropout} with probabilities of 50\% and 20\%, respectively. It is worth mentioning that all convolutional layers and hidden fully connected layers are activated with rectified linear units (ReLU) \cite{Glorot2011AISTATS_ReLU}. Moreover, all the output layers in $\{\textrm{B}_l\}_{l=1}^3$ are activated with the sigmoid function, since all the labels in HCPM are binary.
	
\end{itemize}
The specific configuration of the ETH-CNN structure is presented in Table \ref{tab:cnn}, which lists the numbers of trainable parameters for obtaining different feature maps and vectors.
We can see from this table that there are in total 1,287,189 trainable parameters for ETH-CNN. Thus, the ETH-CNN structure provides a much higher learning capacity, in comparison to the only 1,224 trainable parameters in \cite{Liu16TIP}, which may cause under-fitting issues.
Such abundant trainable parameters also benefit from the extensive training samples of our CPH-Intra database (see Table I-(a) of the \textit{Supporting Document} for the number of training samples).
Another major merit is that all structured labels of HCPM are learned in the ETH-CNN model with shared parameters, to predict ${y}_1(\mathbf{U})$, $\{{y}_2(\mathbf{U}_i)\}_{i=1}^4$ and $\{{y}_3(\mathbf{U}_{i,j})\}_{i,j=1}^4$. This design efficiently reduces the overall computational complexity for predicting the CU partition, compared with conventional learning-based approaches \cite{Hu16BMSB,Liu16DASC,Zhang15TIP,Liu16TIP} that sequentially predict splitting labels from $64\times 64$ CU partition to $16\times 16$ CU partition.

\begin{figure*}
	\centering
	\includegraphics[width=145mm]{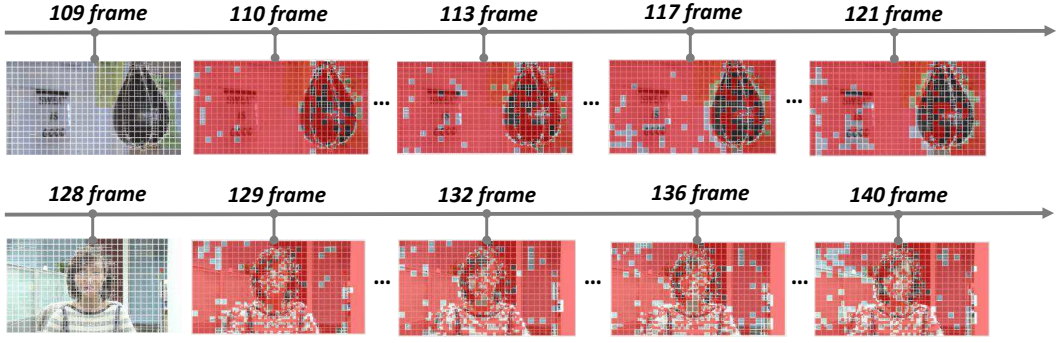}\\
	\caption{\footnotesize{
	    Two examples illustrating temporal CU partition correlation. Red patches represent CUs of the same depths as in the reference frame.
	}}
	\label{fig:cu-depth-subj}
\end{figure*}

\subsection{Loss function for training ETH-CNN model}
\label{sec:training}
Given the above ETH-CNN structure, we concentrate on the loss function for training the ETH-CNN model, which is used to predict HCPM. Assume that there are $R$ training samples, the HCPM labels of which are $\{{y}_1^{r}(\mathbf{U}),\{{y}_2^{r}(\mathbf{U}_i)\}_{i=1}^4$ and $\{{y}_3^{r}(\mathbf{U}_{i,j})\}_{i,j=1}^4\}_{r=1}^{R}$. For each sample, the loss function $L_r$ sums the cross-entropy over all valid elements of HCPM as follows:
\begin{equation} \label{eq:loss-intra}
\begin{aligned}
\begin{split}
L_r=&\:H({y}_1^{r}(\mathbf{U}), \hat{y}_1^{r}(\mathbf{U})) +  \sum_{\substack{i\in\{1,2,3,4\}\\{y}_2^{r}(\mathbf{U}_i)\ne{}\mathrm{null}}} \!\!\!\!\!\! {H({y}_2^{r}(\mathbf{U}_i), \hat{y}_2^{r}(\mathbf{U}_i))}\\& +  \sum_{\substack{i,j\in\{1,2,3,4\}\\{y}_3^{r}(\mathbf{U}_{i,j})\ne{}\mathrm{null}}} \!\!\!\!\!\! {H({y}_3^{r}(\mathbf{U}_{i,j}),\hat{y}_3^{r}(\mathbf{U}_{i,j}))},
\end{split}
\end{aligned}
\end{equation}
where $\{\hat{y}_1^{r}(\mathbf{U}),\{\hat{y}_2^{r}(\mathbf{U}_i)\}_{i=1}^4$ and $\{\hat{y}_3^{r}(\mathbf{U}_{i,j})\}_{i,j=1}^4\}_{r=1}^{R}$ are the labels of HCPM predicted by ETH-CNN. In \eqref{eq:loss-intra}, $H(\cdot,\cdot)$ denotes the cross-entropy between the ground-truth and predicted labels. Considering that some ground-truth labels in HCPM do not exist (such as $\{{y}_3(\mathbf{U}_{2,j})\}_{j=1}^4$ in Fig. \ref{fig:truth-and-pred}), only valid labels with ${y}_2^{r}(\mathbf{U}_i)\ne{}\mathrm{null}$ and $ {y}_3^{r}(\mathbf{U}_{i,j})\ne{}\mathrm{null}$ are counted in the loss of (\ref{eq:loss-intra}).

Then, the ETH-CNN model can be trained by optimizing the loss function over all training samples:
\begin{equation} \label{eq:loss-sum-intra}
\begin{aligned}
L=\frac{1}{R}\sum_{r=1}^R{L_r}.
\end{aligned}
\end{equation}
Because the loss function of \eqref{eq:loss-intra} and \eqref{eq:loss-sum-intra} is the sum of the cross-entropies, the stochastic gradient descent algorithm with momentum is applied to train the ETH-CNN model. Finally, the trained ETH-CNN model can be used to predict the CU partition of HEVC in the form of HCPM.

\subsection{Bi-threshold Decision Scheme}
\label{sec:bi-threshold}
Given an input CTU, ETH-CNN yields the probabilities for the binary labels of $\hat{y}_1(\mathbf{U}), \hat{y}_2(\mathbf{U}_i)$ and $\hat{y}_3(\mathbf{U}_{i,j})$, to predict the CU partition. Let ${P}_1(\mathbf{U}), P_2(\mathbf{U}_i)$ and $P_3(\mathbf{U}_{i,j})$ denote the probabilities of $\hat{y}_1(\mathbf{U})=1, \hat{y}_2(\mathbf{U}_i)=1$ and $\hat{y}_3(\mathbf{U}_{i,j})=1$.  Typically, one decision threshold $\alpha_l$ ($=0.5$ in this paper) is set for each level $l$ of HCPM. Specifically, a CU with ${P}_1(\mathbf{U}) > \alpha_1$ is decided as being split into four CUs $\{\mathbf{U}_i\}_{i=1}^{4}$; otherwise, it is not split. Similarly, $P_2(\mathbf{U}_i) > \alpha_2$ and $P_3(\mathbf{U}_{i,j})> \alpha_3$ make a positive decision on the splitting of $\mathbf{U}_i$ and $\mathbf{U}_{i,j}$. This remarkably reduces the encoding complexity by avoiding all redundant RD cost comparison.

On the other hand, a false prediction of ${y}_1(\mathbf{U})$, ${y}_2(\mathbf{U}_i)$ or ${y}_3(\mathbf{U}_{i,j})$ leads to RD degradation, as the optimal RD performance is not achieved. To ameliorate this issue, we adopt a bi-threshold CU decision scheme \cite{Zhu17TB}. For each level of HCPM, a splitting threshold $\alpha_{l}$ and a non-splitting threshold $\bar{\alpha}_{l}$ are set ($\bar{\alpha}_{l}\leq\alpha_{l}$). Then, the false prediction rate of the ground-truth labels is reduced by introducing an uncertain zone $[\bar{\alpha}_{l}, \alpha_{l}]$, where a parent CU (e.g., $\mathbf{U}, \mathbf{U}_i$ or $\mathbf{U}_{i,j}$) and its four sub-CUs are all checked with the RD cost. Generally speaking, if the uncertain zone becomes wider, more CUs tend to be checked, such that the encoding complexity is increased with decreased misclassification of the CU splitting. Therefore, a trade-off between RD performance and encoding complexity can be achieved through the bi-threshold of $[\bar{\alpha}_{l}, \alpha_{l}]$.

The optimal thresholds of the uncertain zone $[\bar{\alpha}_{l}, \alpha_{l}]$ can be obtained by traversing the combinations of $\{[\bar{\alpha}_{l}, \alpha_{l}]\}_{l=1}^3$ to reduce the encoding complexity as much as possible at the constraint of the RD degradation. However, in practice, both complexity reduction and RD degradation vary in a large extent because of different encoding requirements and diverse video contents. Thus, it is hard to find one specific threshold combination of $\{[\bar{\alpha}_{l}, \alpha_{l}]\}_{l=1}^3$ fitting all encoding requirements and video sequences.
Instead, for the better complexity-RD (CRD) performance, our bi-threshold scheme follows \cite{Zhang15TIP} on account of the setting that the uncertain zones of the CU partition at lower levels should be narrower.
Similar to \cite{Zhang15TIP}, when the RD degradation is constrained in complexity reduction, the uncertain zones at lower levels of the CU partition in our approach are supposed to be narrower, i.e., $[\bar{\alpha}_1, \alpha_1]\subset[\bar{\alpha}_2, \alpha_2]\subset[\bar{\alpha}_3, \alpha_3]$. In addition,the upper and lower thresholds of our bi-threshold scheme are set based on an assumption that the upper and lower thresholds are symmetric with regard to $0.5$, i.e., $\bar{\alpha}_l=1-\alpha_l$ for each level $l$. Note that $0.5$ is a widely used threshold for the conventional binary classification problems.
Given this assumption, the settings of thresholds can be simplified from six variables to three variables for achieving the CRD trade-off.
Moreover, this assumption is reasonable, because the symmetry of the upper and lower thresholds at $0.5$ balances the prediction accuracy for both split and non-split CUs.

\begin{figure}
	\begin{minipage}{0.495\linewidth}
		\centering
		\includegraphics[width=36mm]{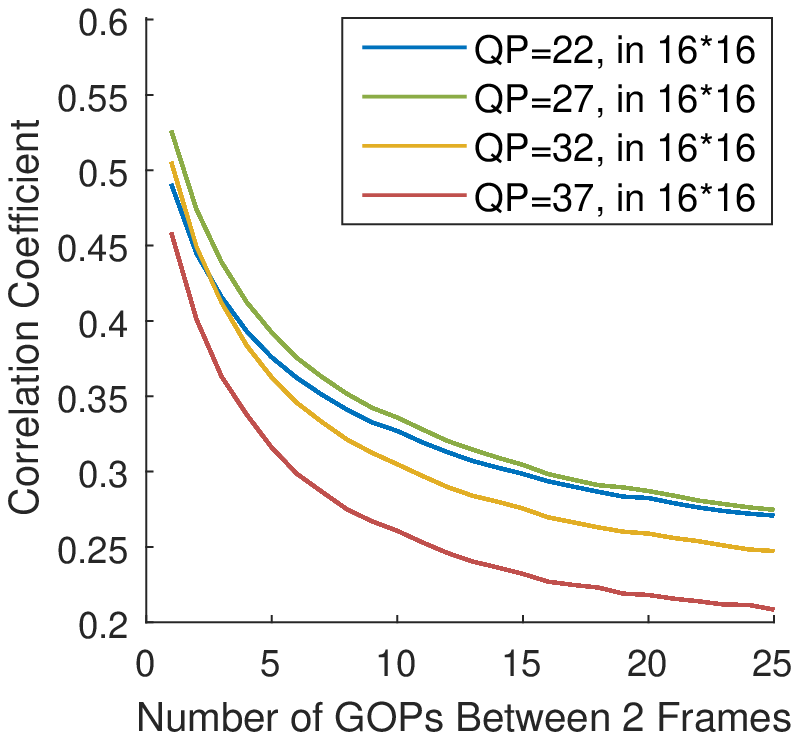}\\
		\vspace{0.9em}
		%\centerline{\footnotesize{(a)}}\medskip
	\end{minipage}
	\hfill
	\begin{minipage}{0.495\linewidth}
		\centering
		\includegraphics[width=36mm]{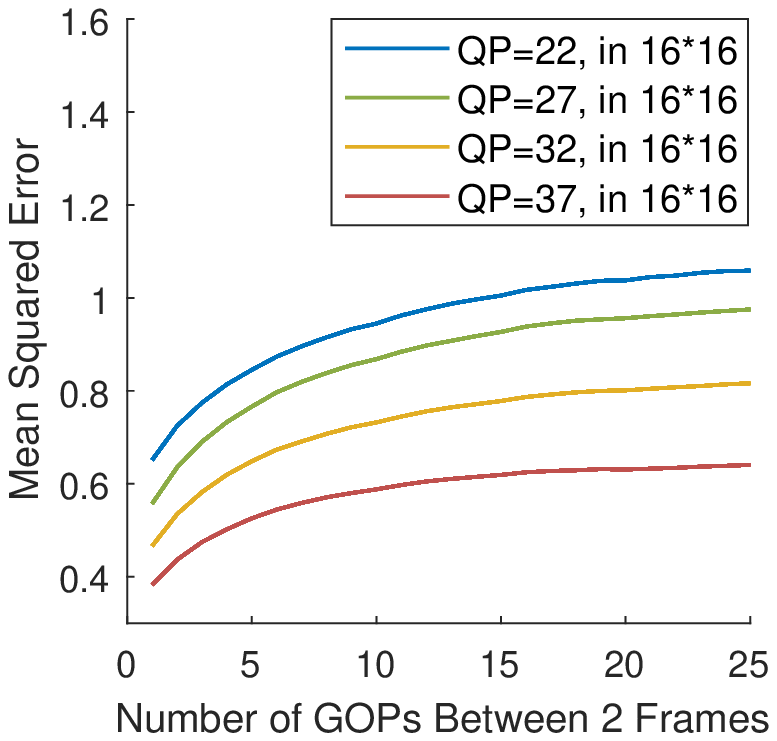}\\
		\vspace{0.9em}
		%\centerline{\footnotesize{(b)}}\medskip
	\end{minipage}
	\vfill
	\begin{minipage}{0.495\linewidth}
		\centering
		\includegraphics[width=36mm]{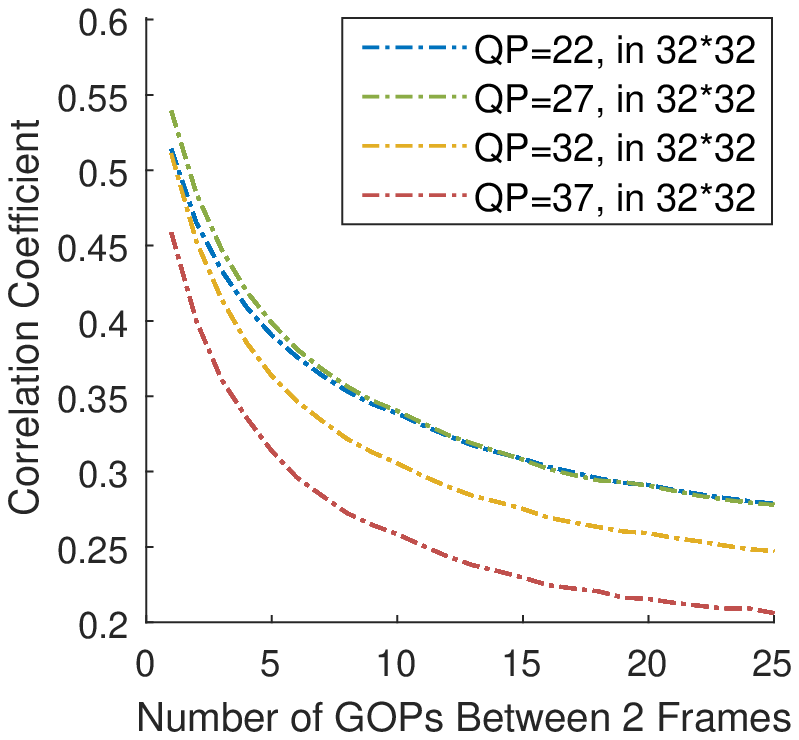}\\
		\vspace{0.9em}
		%\centerline{\footnotesize{(c)}}\medskip
	\end{minipage}
	\hfill
	\begin{minipage}{0.495\linewidth}
		\centering
		\includegraphics[width=36mm]{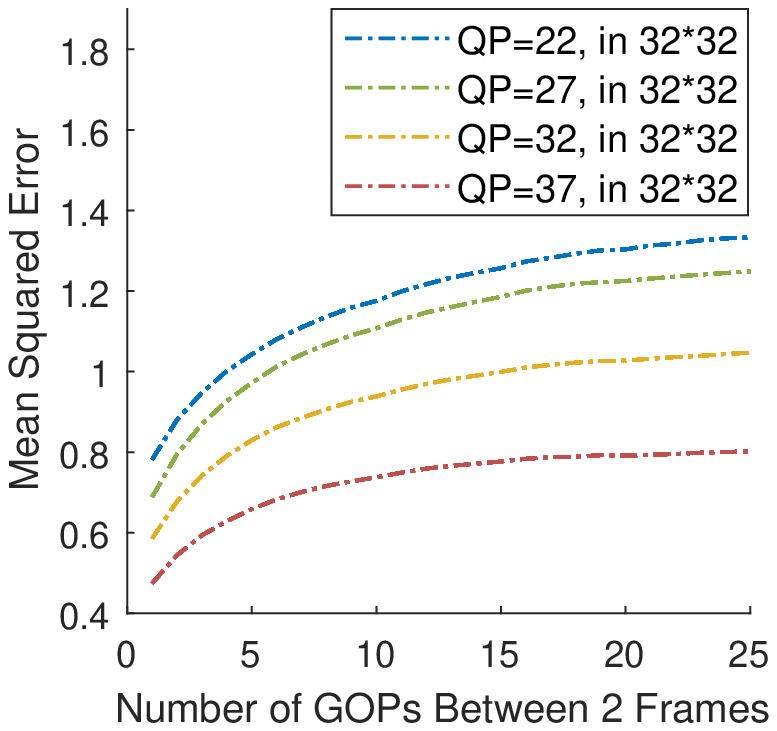}\\
		\vspace{0.9em}
		%\centerline{\footnotesize{(d)}}\medskip
	\end{minipage}
	\vfill
	\begin{minipage}{0.495\linewidth}
		\centering
		\includegraphics[width=36mm]{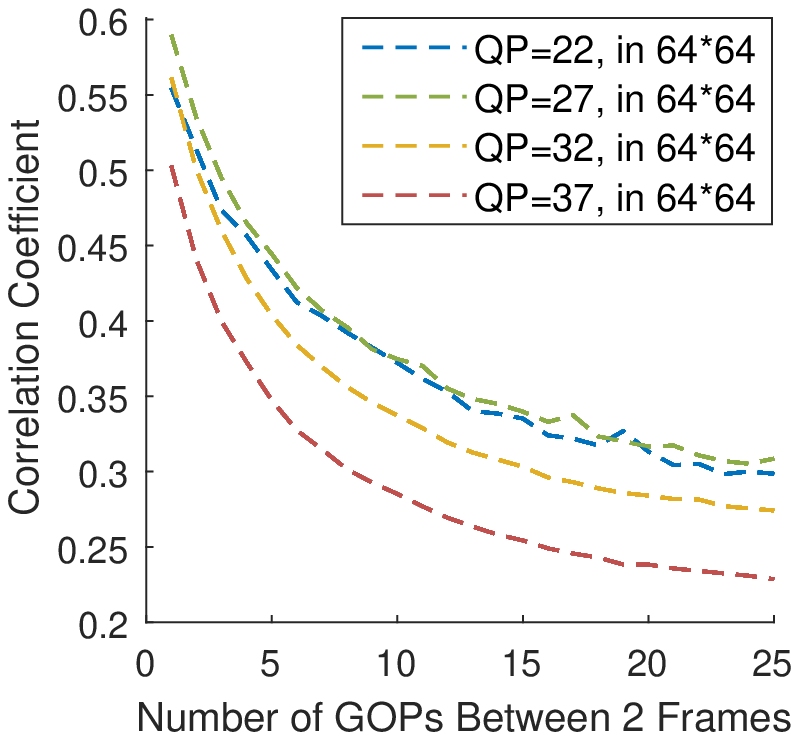}\\
		%\centerline{\footnotesize{(e)}}\medskip
	\end{minipage}
	\hfill
	\begin{minipage}{0.495\linewidth}
		\centering
		\includegraphics[width=36mm]{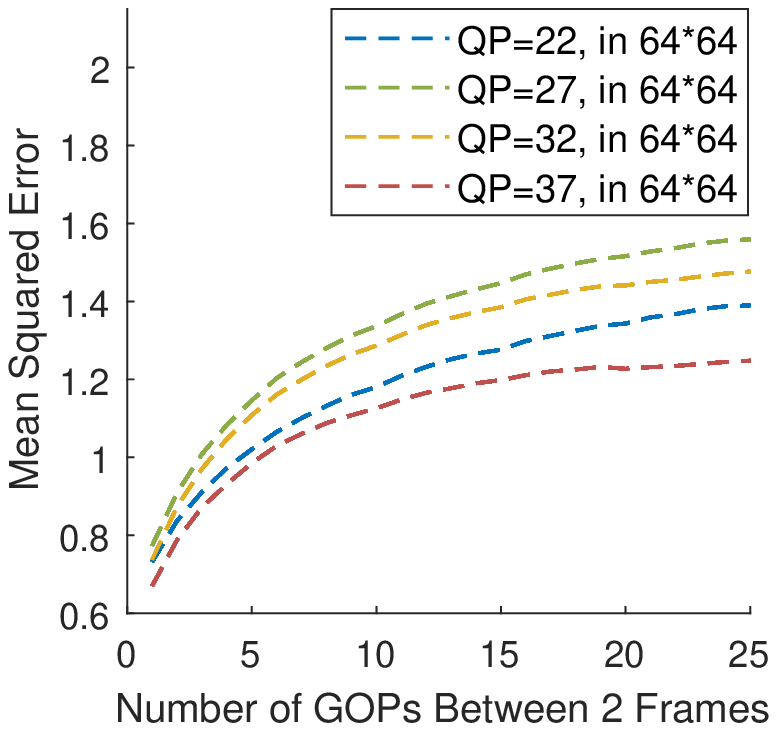}\\
		%\centerline{\footnotesize{(f)}}\medskip
	\end{minipage}
	\caption{\footnotesize{
			Temporal CU depth correlation versus distances of two frames for inter-mode HEVC at four QP values. Left: Correlation coefficient. Right: Mean squared error.
	}}
	\label{fig:cu-depth-obj}
\end{figure}

\subsection{Computational Complexity of ETH-CNN}
\label{sec:complexity-cnn}
In this section, we analyze the time complexity of ETH-CNN via counting the number of floating-point operations, including additions and multiplications. Note that all floating-point operations in ETH-CNN are in single precision (32-bit). The last two columns of Table \ref{tab:cnn} report the numbers of floating-point operations for each layer of ETH-CNN and the total number of floating-point operations required in ETH-CNN. We can see that ETH-CNN performs a total of $3.05\times 10^6$ floating-point operations, including 1,497,584 additions and 1,552,149 multiplications. ETH-CNN performs fewer floating-point operations than Alex-Net ($\sim 3\times 10^9$ floating-point operations) or VGG-Net ($\sim 4\times 10^{10}$ floating-point operations) by at least three orders of magnitude.

In Table \ref{tab:cnn}, early termination is not considered for counting the total number of floating-point operations of ETH-CNN. In practical compression, the early termination of ETH-CNN can further reduce the time complexity. Here, we record the average number of floating-point operations of ETH-CNN, when compressing 18 standard JCT-VC sequences \cite{Ohm12TCSVT} at intra-mode and QP = 32. We find that $12.6 \%$ floating-point operations can be saved due to the early termination mechanism in ETH-CNN.
Consequently, for compressing 1080p@30 Hz sequences, our approach requires $\sim$ 40.8 G single-precision floating-point operations per second (FLOPS) for intra-mode HEVC complexity reduction. The FLOPS of our approach is far less than the computational capacity of an Intel(R) Core(TM) i7-7700K CPU @4.2 GHz, which supports $\sim$ 287.9 G double-precision FLOPS, as reported in \cite{IntelCPU2017}.
In the \textit{Supporting Document}, time complexity of our approach on field programmable gate array (FPGA) implementation is also analyzed.
As analyzed, ETH-CNN only consumes less than 3 ms for one frame of 1080p sequences in the FPGA of Virtex UltraScale+ VU13P.

\section{Complexity Reduction for Inter-mode HEVC}
\label{sec:method-inter}
In this section, we first analyze the temporal correlation of the CU depth in inter-mode HEVC. Based on our analysis, we then propose to predict the CU partition of inter-mode HEVC, via designing the ETH-LSTM network. Finally, the computational complexity of ETH-LSTM is analyzed.

\subsection{Analysis on Temporal Correlation of CU Depth}
\label{sec:analysis-inter}
Typically, the adjacent video frames exhibit similarity in video content, and such similarity decays as the temporal distance of two adjacent frames increases. The same phenomenon may also hold for the CU partition of inter-mode HEVC. Fig. \ref{fig:cu-depth-subj} shows examples of the CU partition across adjacent frames with different temporal distances. In this figure, we consider the first column as the reference. Then, the CU partition of subsequent frames with different distances is compared to the reference. In each CU, we highlight the same CU depth in red. We can see from Fig. \ref{fig:cu-depth-subj} that there exists high similarity in the CU partition across neighboring frames, and the similarity drops along with increasing temporal distance.

We further measure the similarity of the CU partition in inter-mode HEVC over all 93 sequences in the training and validation sets of our CPH-Inter database. Specifically, we calculate the correlation of the CU depth between two frames at various distances, ranging from 1 group of pictures (GOP) to 25 GOPs. Such correlation is measured between co-located units\footnote{Non-overlapping units with sizes of $64\times 64$, $32 \times 32$ and $16\times 16$ are considered, corresponding to splitting depths of 0, 1 and 2. Note that $8 \times 8$ units are not measured, as an $8 \times 8$ CU is definitely split from a larger CU.} from two frames in terms of the linear correlation coefficient (CC) and mean squared error (MSE). In our analysis, the results for the CC and MSE are averaged over all frames of 93 sequences, which are shown in Fig. \ref{fig:cu-depth-obj} for four QPs (QP = 22, 27, 32 and 37).

We can see from Fig. \ref{fig:cu-depth-obj} that CC is always much larger than 0, indicating the existence of a positive correlation on the temporal CU depth. Moreover, the CC decreases alongside increasing distance between two frames. Similar results hold for MSE, as can be found in Fig. \ref{fig:cu-depth-obj}. Therefore, Fig. \ref{fig:cu-depth-obj} implies that there exist long- and short-term dependencies of the CU partition across adjacent frames for inter-mode HEVC.

\subsection{Deep LSTM Structure}	
\label{sec:lstm_struct}

\begin{figure}
	\centering
	\includegraphics[width=70mm]{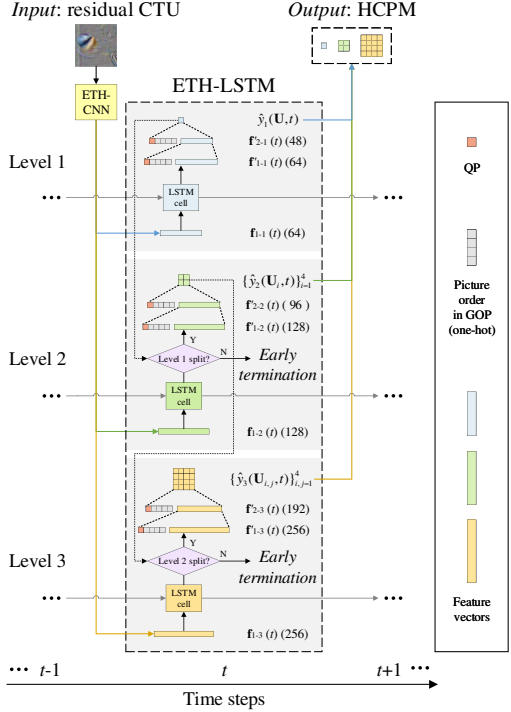}\\
	\caption{\footnotesize{
			Framework of ETH-LSTM. The number in each bracket is the dimension of the vector.
	}}
	\label{fig:lstm}
\end{figure}

As analyzed in Section \ref{sec:analysis-inter}, the CU partition of neighboring frames is correlated with each other. Thus, this section proposes the ETH-LSTM approach, which learns the long- and short-term dependencies of the CU partition across frames. Fig. \ref{fig:lstm} illustrates the overall framework of ETH-LSTM.
As observed in Fig. \ref{fig:lstm}, the input to ETH-LSTM is the residue of each CTU. Here, the residue is obtained by precoding the currently processed frame, in which both the CU and PU sizes are forced to be 64$\times$64. It is worth mentioning that the computational time for precoding consumes less than $3\%$ of the total encoding time, which is considered in our approach as time overhead. Then, the residual CTU is fed into ETH-CNN. For inter-frames, the parameters of ETH-CNN are re-trained over the residue and ground-truth splitting of the training CTUs from the CPH-Inter database. Next, the features $\{\mathbf{f}_{1-l}\}_{l=1}^3$ of ETH-CNN are extracted for frame $t$, and these features are then fed into ETH-LSTM. Recall that $\{\mathbf{f}_{1-l}\}_{l=1}^3$ are the features at Layer 7 of ETH-CNN.

Fig. \ref{fig:lstm} shows that three LSTMs in ETH-LSTM are arranged in a hierarchy for determining the CU depths in the form of HCPM. Specifically, the LSTM cells are at levels 1, 2 and 3, corresponding to three levels of HCPM: $\hat{y}_1(\mathbf{U},t)$, $\{\hat{y}_2(\mathbf{U}_i,t)\}_{i=1}^4$ and $\{\hat{y}_3(\mathbf{U}_{i,j},t)\}_{i,j=1}^4$. Here, $\hat{y}_1(\mathbf{U},t)$ indicates whether the CU $\mathbf{U}$ (size: $64\times 64$, depth = 0 ) at frame $t$ is split. Similarly, $\{\hat{y}_2(\mathbf{U}_i,t)\}_{i=1}^4$ and $\{\hat{y}_3(\mathbf{U}_{i,j},t)\}_{i,j=1}^4$ denote the splitting labels of CUs $\mathbf{U}_i$ (size: $32\times 32$, depth = 1) and $\mathbf{U}_{i,j}$ (size: $16\times 16$, depth = 2). At each level, two fully connected layers follow the LSTM cells, which also include the QP value and the order of frame $t$ at GOP. Note that the frame order is represented by a one-hot vector. For level $l$ at frame $t$, we denote $\mathbf{f'}_{1-l}(t)$ and $\mathbf{f'}_{2-l}(t)$ as the output features of the LSTM cell and the first fully connected layer. Additionally, the output of the second fully connected layer is the probabilities of CU splitting, which are binarized to predict HCPM\footnote{The binarization process is the same as that of ETH-CNN in Section \ref{sec:cnn_struct}. The bi-threshold decision scheme of Section \ref{sec:bi-threshold} can also be applied for the binarization process to achieve a trade-off between prediction accuracy and complexity reduction.}. However, if the LSTM cell at level 1 decides not to split, the calculation of $\{\hat{y}_2(\mathbf{U}_i,t)\}_{i=1}^4$ of HCPM is terminated early at level 2, via skipping its two fully connected layers.
Instead, all indices of $\{\hat{y}_2(\mathbf{U}_i,t)\}_{i=1}^4$ are set to be 0, meaning that the CU depths at level 2 are 0. The calculation of  $\{\hat{y}_3(\mathbf{U}_{i,j},t)\}_{i,j=1}^4$ at level 3 is also terminated early in a similar manner. Consequently, early termination is able to reduce the computational time of ETH-LSTM. Finally, the results of HCPM are output by ETH-LSTM, therein predicting the partition patterns of a CTU at frame $t$.

When determining the HCPM of each CTU, the partition patterns of the CTUs co-located at the previous frames can be considered in ETH-LSTM. ETH-LSTM learns the long- and short-term correlations of CU depths across frames, via incorporating the LSTM cells at different levels.
For learning ETH-LSTM, each LSTM cell is trained separately using its corresponding CUs.
Next, we introduce the learning mechanism of ETH-LSTM by taking the LSTM cell of frame $t$ at level $l$ as  an example. Such an LSTM cell works with three gates: the input gate $\mathbf{i}_{l}(t)$, the output gate $\mathbf{o}_{l}(t)$ and the forget gate $\mathbf{g}_{l}(t)$. Given the input ETH-CNN feature $\mathbf{f}_{1-l}(t)$ and the output feature $\mathbf{f'}_{1-l}(t-1)$ of the LSTM cell at the last frame, these three gates can be obtained by
\begin{equation} \label{eq:LSTM-gate}
\left\{
\begin{array}{lcl}
\mathbf{i}_{l}(t)=\sigma(\mathbf{W}_i\cdot[\mathbf{f}_{1-l}(t),\mathbf{f'}_{1-l}(t-1)]+\mathbf{b}_i)\\
\mathbf{o}_{l}(t)=\sigma(\mathbf{W}_o\cdot[\mathbf{f}_{1-l}(t),\mathbf{f'}_{1-l}(t-1)]+\mathbf{b}_o)\\
\mathbf{g}_{l}(t)=\sigma(\mathbf{W}_f\cdot[\mathbf{f}_{1-l}(t),\mathbf{f'}_{1-l}(t-1)]+\mathbf{b}_f)\\
\end{array}\right.
\end{equation}
where $\sigma(\cdot)$ stands for the sigmoid function. In the above equations, $\mathbf{W}_i$, $\mathbf{W}_o$, $\mathbf{W}_f$ are trainable parameters for three gates, and $\mathbf{b}_i$, $\mathbf{b}_o$ and $\mathbf{b}_f$ are their biases. With these three gates, the LSTM cell updates its state at frame $t$ as
\begin{equation} \label{eq:LSTM-state}
\begin{aligned}
%\mathbf{c}_{l}(t)=\mathbf{g}_{l}(t)\odot \mathbf{c}_{l}(t-1)+\mathbf{i}_{l}(t)\circ \tanh(\mathbf{W}_c\odot[\mathbf{f}_{1-l}(t),\mathbf{f'}_{1-l}(t-1)]+\mathbf{b}_c),
\begin{split}
\mathbf{c}_{l}(t)=\:&\mathbf{i}_{l}(t)\odot \tanh(\mathbf{W}_c\odot[\mathbf{f}_{1-l}(t),\mathbf{f'}_{1-l}(t-1)]+\mathbf{b}_c)\\
&\!\!+\mathbf{g}_{l}(t)\odot \mathbf{c}_{l}(t-1),
\end{split}
\end{aligned}
\end{equation}
where $\odot$ represents element-wise multiplication. In \eqref{eq:LSTM-state}, $\mathbf{W}_c$ and $\mathbf{b}_c$ are parameters and biases for $\mathbf{c}_{l}(t)$, which need to be trained. Finally, the output of the LSTM cell $\mathbf{f'}_{1-l}(t)$ can be calculated as follows:
\begin{equation} \label{eq:LSTM-output}
\begin{aligned}
\mathbf{f'}_{1-l}(t)=\mathbf{o}_{l}(t)\odot\mathbf{c}_{l}(t).
\end{aligned}
\end{equation}
Note that both $\mathbf{c}_{l}(t)$ and $\mathbf{f'}_{1-l}(t)$ are vectors with the same length as $\mathbf{f}_{1-l}(t)$.

The configuration of ETH-LSTM with all trainable parameters is listed in Table \ref{tab:lstm}. % This sentence is added on 20170806.
For training the parameters of the LSTM cells in the above equations, the cross-entropy is applied as the loss function, the same as that for ETH-CNN defined in \eqref{eq:loss-intra}. Let $L_r(t)$ be the loss for the $r$-th training CU sample at frame $t$. Then, each LSTM cell among the three levels of ETH-LSTM can be trained by optimizing the loss over all $R$ training samples alongside $T$ frames,
\begin{equation} \label{eq:loss-sum-inter}
\begin{aligned}
L=\frac{1}{RT}\sum_{r=1}^R\sum_{t=1}^{T}{L_r(t)},
\end{aligned}
\end{equation}
which is solved by the stochastic gradient descent algorithm with momentum. Finally, given the trained LSTM cells, ETH-LSTM can be used to generate HCPM for predicting the CU partition at the inter-frames.

\begin{table}
	\scriptsize
	\newcommand{\tabincell}[2]{\begin{tabular}{@{}#1@{}}#2\end{tabular}}
	\begin{center}
		\caption{\footnotesize{Configuration of ETH-LSTM}}
		\label{tab:lstm}
		\begin{tabular}{|c|c|c|c|c|}
			\hline Feature & \tabincell{c}{Size} & \tabincell{c}{Number of\\ Parameters} & \tabincell{c}{Number of\\Additions} & \tabincell{c}{Number of\\Multiplications} \\
			\hline $\mathbf{i}_{1}$/$\mathbf{o}_{1}$/$\mathbf{g}_{1}(t)$  & 64 & 8,192 & 8,128 & 8,192\\
			\hline $\mathbf{i}_{2}$/$\mathbf{o}_{2}$/$\mathbf{g}_{2}(t)$ & 128 & 32,768 & 32,640 & 32,768\\
			\hline $\mathbf{i}_{3}$/$\mathbf{o}_{3}$/$\mathbf{g}_{3}(t)$& 256 & 131,072 & 130,816 & 131,072\\
			\hline $\mathbf{c}_{1}(t)$ & 64 & 8,192 & 8,255 & 8,320\\
			\hline $\mathbf{c}_{2}(t)$ & 128 & 32,768 & 32,895 & 33,024 \\
			\hline $\mathbf{c}_{3}(t)$ & 256 & 131,072 & 131,327 & 131,584\\
			\hline $\mathbf{f}'_{1-1}(t)$ & 64 & 0 & 63 & 64 \\
			\hline $\mathbf{f}'_{1-2}(t)$ & 128 & 0 & 127 & 128 \\
			\hline $\mathbf{f}'_{1-3}(t)$ & 256 & 0 & 255 & 256 \\
			\hline $\mathbf{f}'_{2-1}(t)$ & 48 & 3,312 & 3,264 & 3,312\\
			\hline $\mathbf{f}'_{2-2}(t)$ & 96 & 12,768 & 12,672 & 12,768\\
			\hline $\mathbf{f}'_{2-3}(t)$ & 192 & 50,112 & 49,920 & 50,112\\
			\hline $\hat{y}_1(\mathbf{U},t)$ & 1 & 53 & 52 & 53\\
			\hline $\{\hat{y}_2(\mathbf{U}_i,t)\}_{i=1}^4$ & 4 & 404 & 400 & 404\\
			\hline $\{\hat{y}_3(\mathbf{U}_{i,j},t)\}_{i,j=1}^4$ & 16 & 3,152 & 3,136 & 3,152\\
			\hline Total & - & 757,929 & 757,118 & 759,273 \\
			\hline
		\end{tabular}
	\end{center}
\end{table}

\subsection{Computational Complexity of ETH-LSTM}
\label{sec:complexity-lstm}

As shown in the last two columns of Table \ref{tab:lstm}, there are in total 757,118 additions and 759,273 multiplications for ETH-LSTM, almost half of those for ETH-CNN.
With the early termination, the total number of floating-point operations in Table  \ref{tab:lstm} can be further reduced by $4.3\%$ when running ETH-LSTM for compressing 18 standard JCT-VC sequences at QP = 32 (the inter-mode configuration is the same as that in the section of experimental results ). Because ETH-LSTM requires the extracted features of ETH-CNN as the input, our complexity reduction approach consumes in total $4.37 \times 10^6$ single-precision floating-point operations for a CTU.
In other words, our approach consumes 66.9 G single-precision FLOPS for 1080p@30 Hz sequences, much less than the 287.9 G double-precision FLOPS for an Intel(R) Core(TM) i7-7700K CPU @4.2 GHz. As analyzed in the \textit{Supporting Document}, our approach requires less than 3 ms/frame running time in the FPGA device of Virtex UltraScale+ VU13P.

%As shown in Table \ref{tab:lstm}, there exist 757,118 additions and 759,273 multiplications in ETH-LSTM, only around one half of those in ETH-CNN, remaining succinct. The main reason of the limited numbers lies in the short input vectors $\{\mathbf{f}_{1-l}(t)\}_{l=1}^3$ of ETH-LSTM, with merely 64, 128 and 256 high-level features extracted from ETH-CNN at three levels, which are still sufficient to predict inter-mode CU partition.
%With early termination, the numbers of floating-point operations in ETH-LSTM can be reduced by $2\%\sim 5\%$ at QPs $\in\{22,27,32,37\}$, illustrated in Fig. \ref{fig:oper-times} (b).
%Combining both ETH-CNN and ETH-LSTM, 1.10G additions and 1.13G multiplications are needed to encode a 1080p frame at QP 32, also suitable for a mainstream desktop CPU \cite{IntelCPU2017}.

\section{Experimental Results}
\label{sec:result}

In this section, we present experimental results to validate the effectiveness of our approach in reducing HEVC complexity at both intra- and inter-modes. For evaluating performance at intra-mode, we compare our approach with two state-of-the-art approaches: the SVM based approach \cite{Zhang15TIP} and the CNN based approach \cite{Liu16TIP}. For evaluating performance at inter-mode, we compare our approach to the latest approaches of \cite{Zhang15TIP}, \cite{Correa15TCSVT} and \cite{Mallikarachchi16TCSVT}. In Section \ref{sec:CandS}, we discuss the configuration of the experiments and the settings of our approach.  In Sections \ref{sec:PEintra} and \ref{sec:PEiner}, we compare the performance of our and other approaches for intra- and inter-mode HEVC, respectively. Finally, Section \ref{sec:runningtime} analyzes the running time of our complexity reduction approach.

\subsection{Configuration and Settings}
\label{sec:CandS}

\textbf{Configuration of experiments.} In our experiments, all complexity reduction approaches were implemented in the HEVC reference software HM 16.5 \cite{HM}. In HM 16.5, the AI configuration was applied with the default configuration file \emph{encoder\_intra\_main.cfg}\cite{TestCond13} for the performance evaluation at intra-mode, and the LDP configuration was used with the default configuration file \emph{encoder\_low\_delay\_P\_main.cfg}\cite{TestCond13} for the performance evaluation at inter-mode. Here, four QP values, $\{22,27,32,37\}$, were chosen to compress the images and video sequences.
At both intra- and inter-modes, our experiments were tested on 18 video sequences of the JCT-VC standard test set \cite{Ohm12TCSVT}. Moreover, all 200 test images from the CPH-Intra database were evaluated for intra-mode HEVC complexity reduction.
In our experiments, the Bj\o{}ntegaard delta bit-rate (BD-BR) and Bj\o{}ntegaard delta PSNR (BD-PSNR) \cite{Bjontegaard2011} were measured to assess the RD performance. Additionally, $\Delta{}T$, which denotes the encoding time-saving rate over the original HM, was used to measure the complexity reduction.
All experiments were conducted on a computer with an Intel(R) Core (TM) i7-7700K CPU @4.2 GHz, 16 GB RAM and the Windows 10 Enterprise 64-bit operating system.
Note that a GeForce GTX 1080 GPU was used to accelerate the training speed, but it was disabled when testing the HEVC complexity reduction.

\textbf{Training Settings.}
In our complexity reduction approach, two ETH-CNN models were trained from the training sets of the CPH-Intra and CPH-Inter databases, corresponding to intra- and inter-mode HEVC, respectively. In addition, one ETH-LSTM model was trained from the training set of the CPH-Inter database. In training these models, the hyper-parameters were tuned on the validation sets of the CPH-Intra and CPH-Inter databases. Specifically, all trainable parameters in ETH-CNN and ETH-LSTM were randomly initialized, obeying the truncated normal distribution with zero mean and standard deviation of 0.1. The batch size $R$ for training was 64, and the momentum of the stochastic gradient descent algorithm was set to 0.9. To train the ETH-CNN models, the initial learning rate was 0.01 and decreased by $1\%$ exponentially every 2,000 iterations, and there were in total 1,000,000 iterations. To train ETH-LSTM, the initial learning rate was 0.1 and decreased by $1\%$ exponentially every 200 iterations, and the total number of iterations was 200,000. The length of the trained ETH-LSTM model was $T=20$. Moreover, at the LDP configuration, the overlapping of 10 frames was applied to generate the training samples of the ETH-LSTM model. Such overlapping was introduced to enhance the diversity of the training samples and meanwhile augment the training data.

\textbf{Test Settings.}
For the test, $T$ was set equal to the number of P frames in the sequence, where the internal information of ETH-LSTM can propagate throughout the whole sequence to leverage the long- and short-term dependencies. Note that ETH-LSTM is invoked step-wise during the test, i.e., the hidden state and the output of frame $t$ are calculated after the processing of frame $t-1$ is accomplished, instead of yielding the output of all $T$ frames at one time. For the bi-threshold decision scheme, at intra-mode the values of the thresholds were $\bar{\alpha}_l=\alpha_l=0.5$ where $l\in\{1,2,3\}$ is the CU depth. It is because our approach is able to achieve sufficiently small loss of RD performance at intra-mode and the bi-threshold decision is not necessary. At inter-mode, we set $[\bar{\alpha}_1,\alpha_1]=[0.4, 0.6]$, $[\bar{\alpha}_2,\alpha_2]=[0.3, 0.7]$ and $[\bar{\alpha}_3,\alpha_3]=[0.2, 0.8]$, corresponding to the different CU depth of $l\in\{1,2,3\}$. Here, we followed \cite{Zhang15TIP} to apply this bi-threshold setting, which can efficiently enhance RD performance with desirable rate of complexity reduction. The width of uncertain zone increases along with increased level $l$ of CU depth, since the smaller CU depth tends to early bypass redundant checking of CUs in RD optimization (RDO), which is achieved by a narrower uncertain zone $[\bar{\alpha}_l,\alpha_l]$.

%
%Considering that each residual frame is not conclusive until its previous frames are all encoded by our approach, thus the step-wise invoking ensures the timeliness of the residual frame, essential for ETH-LSTM to achieve desirable predicting accuracy.

\subsection{Performance Evaluation at Intra-mode}
\label{sec:PEintra}

\begin{figure}
	\centering
	\begin{minipage}{0.49\linewidth}
		\centering
		\includegraphics[width=45mm]{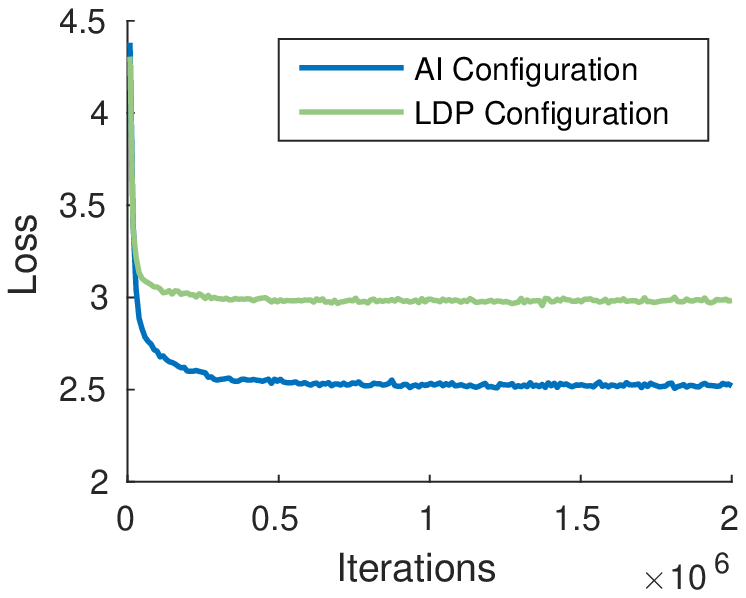}\\
		\centerline{\footnotesize{(a) ETH-CNN}}\medskip
	\end{minipage}
	\hfill
	\begin{minipage}{0.49\linewidth}
		\centering
		\includegraphics[width=45mm]{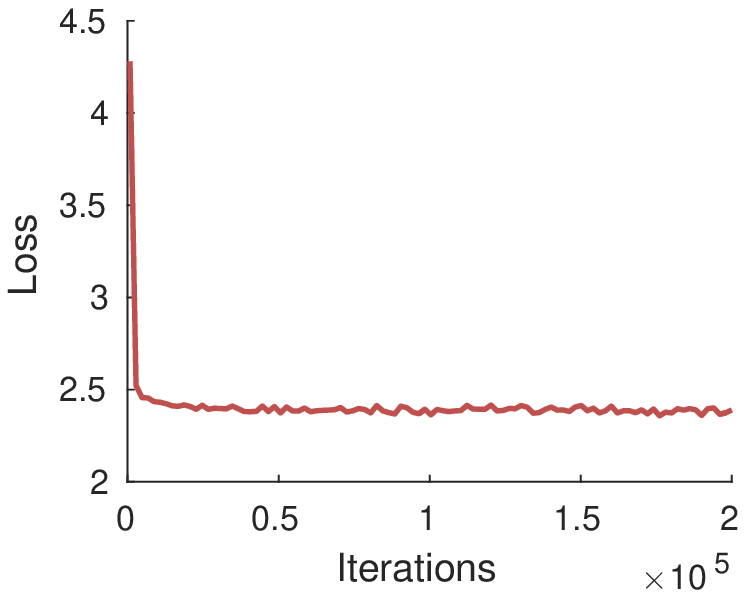}\\
		\centerline{\footnotesize{(b) ETH-LSTM}}\medskip
	\end{minipage}
	\caption{Training loss alongside iterations when learning the models of ETH-CNN at intra-mode, ETH-CNN at inter-mode and ETH-LSTM. Note that the training loss of intra-mode ETH-CNN is over the training set of CPH-Intra database, while the training loss of inter-mode ETH-CNN and ETH-LSTM is over the training set of CPH-Inter database.}\label{fig:train-loss-accu}
\end{figure}

\begin{table}
	\tiny
	\newcommand{\tabincell}[2]{\begin{tabular}{@{}#1@{}}#2\end{tabular}}
	\begin{center}
		\caption{\footnotesize{Results for images from our CPH-Intra test set (AI)}}	
		\label{tab:result-image-intra}
		\begin{tabular}{|c|c|c|c|c|c|c|c|c|}
			\hline \multirow{2}{*}{\tabincell{c}{Image\\Source}} & \multirow{2}{*}{Resolution} & \multirow{2}{*}{Appr.} & \multirow{2}{*}{\tabincell{c}{\hspace{-0.5em}BD-BR\hspace{-0.5em} \\(\%)}} & \multirow{2}{*}{\tabincell{c}{\hspace{-0.5em}BD-PSNR\hspace{-0.5em} \\(dB)}} & \multicolumn{4}{|c|}{$\Delta{}T$ (\%)} \\
			\cline{6-9} & & & & & QP=22 & QP=27 & QP=32 & QP=37 \\
			
			\hline \multirow{12}{*}{\tabincell{c}{CPH-Intra\\Test Set}} & \multirow{3}{*}{768$\times$512} & \cite{Liu16DASC} & 5.113 & -0.343 & \textbf{-59.43} & -54.70 & -48.74 & -44.83 \\
			& & \cite{Liu16TIP} & 2.885 & -0.210 & -54.97 & \textbf{-58.78} & \textbf{-61.78} & \textbf{-64.41} \\
			& & Our & \textbf{0.935} & \textbf{-0.067} & -51.10 & -55.16 & -57.31 & -62.18 \\
			
			\cline {2-9} & \multirow{3}{*}{1536$\times$1024} & \cite{Liu16DASC} & 6.002 & -0.374 & -58.94 & -54.85 & -50.57 & -50.95  \\
			& & \cite{Liu16TIP} & 3.134 & -0.208 & -55.84 & -59.46 & -62.43 & -64.17 \\
			& & Our & \textbf{1.220} & \textbf{-0.081} & \textbf{-59.50} & \textbf{-60.78} & \textbf{-64.71} & \textbf{-66.76} \\
			
			\cline {2-9} & \multirow{3}{*}{2880$\times$1920} & \cite{Liu16DASC} & 4.035 & -0.207 & -57.03 & -52.79 & -52.31 & -59.51 \\
			& & \cite{Liu16TIP} & 2.130 & -0.115 & -59.95 & -63.14 & -68.07 & -69.46 \\
			& & Our & \textbf{1.661} & \textbf{-0.094} & \textbf{-70.91} & \textbf{-74.48} & \textbf{-76.06} & \textbf{-78.47} \\
			
			\cline {2-9} & \multirow{3}{*}{4928$\times$3264} & \cite{Liu16DASC} & 4.630 & -0.209 & -58.02 & -62.74 & -65.30 & -67.46\\
			& & \cite{Liu16TIP} & 1.863 & -0.086 & -61.43 & -65.27 & -68.70 & -71.00 \\
			& & Our & \textbf{1.728} & \textbf{-0.082} & \textbf{-75.51} & \textbf{-76.96} & \textbf{-78.34} & \textbf{-78.68} \\
			
			\hline \multicolumn{2}{|c|}{\multirow{3}{*}{Std. dev.}} & \cite{Liu16DASC} & 0.720 & 0.076 & \textbf{0.92} & 3.82 & 6.51 & 8.57 \\
			\multicolumn{2}{|c|}{} & \cite{Liu16TIP} & 0.604 & 0.064 & 3.13 & \textbf{3.07} & \textbf{3.64} & \textbf{3.48} \\
			\multicolumn{2}{|c|}{} & Our & \textbf{0.376} & \textbf{0.011} & 11.05 & 10.55 & 9.87 & 8.36 \\
			
			\hline \multicolumn{2}{|c|}{\multirow{3}{*}{Best}} & \cite{Liu16DASC} & 4.035 & -0.207 & -59.43 & -62.74 & -65.30 & -67.46 \\
			\multicolumn{2}{|c|}{} & \cite{Liu16TIP} & 1.863 & -0.086 & -61.43 & -65.27 & -68.70 & -71.00 \\
			\multicolumn{2}{|c|}{} & Our & \textbf{0.935} & \textbf{-0.067} & \textbf{-75.51} & \textbf{-76.96} & \textbf{-78.34} & \textbf{-78.68} \\
			
			\hline \multicolumn{2}{|c|}{\multirow{3}{*}{Average}} & \cite{Liu16DASC} & 4.945 & -0.284 & -58.36 & -56.27 & -54.23 & -55.69 \\
			\multicolumn{2}{|c|}{} & \cite{Liu16TIP} & 2.353 & -0.155 & -58.05 & -61.66 & -65.25 & -67.26 \\
			\multicolumn{2}{|c|}{} & Our & \textbf{1.386} & \textbf{-0.081} & \textbf{-64.01} & \textbf{-66.09} & \textbf{-68.11} & \textbf{-70.52} \\
			\hline
			
		\end{tabular}
	\end{center}
\end{table}

\begin{table}
	\tiny
	\newcommand{\tabincell}[2]{\begin{tabular}{@{}#1@{}}#2\end{tabular}}
	\begin{center}
		\caption{\footnotesize{Results for sequences of the JCT-VC test set (AI)}}
		\label{tab:result-video-intra}
		\begin{tabular}{|c|c|c|c|c|c|c|c|c|}
			\hline \multirow{2}{*}{Class} & \multirow{2}{*}{Sequence} & \multirow{2}{*}{Appr.} & \multirow{2}{*}{\tabincell{c}{\hspace{-0.5em}BD-BR\hspace{-0.5em} \\(\%)}} & \multirow{2}{*}{\tabincell{c}{\hspace{-0.5em}BD-PSNR\hspace{-0.5em} \\(dB)}} & \multicolumn{4}{|c|}{$\Delta{}T$ (\%)} \\
			\cline{6-9} & & & & & QP=22 & QP=27 & QP=32 & QP=37 \\
			
			\hline \multirow{6}{*}{A} & \multirow{3}{*}{\textit{PeopleOnStreet}} & \cite{Liu16DASC} & 9.627 & -0.492 & -52.12 & -50.63 & -37.79 & -34.81 \\
			& & \cite{Liu16TIP} & 3.969 & -0.209 & -50.79 & -53.87 & -56.58 & -61.15 \\
			& & Our & \textbf{2.366} & \textbf{-0.126} & \textbf{-59.16} & \textbf{-59.42} & \textbf{-61.63} & \textbf{-63.82} \\
			\cline {2-9} & \multirow{3}{*}{\textit{Traffic}} & \cite{Liu16DASC} & 6.411 & -0.304 & -37.11 & -25.36 & -19.63 & -33.38 \\
			& & \cite{Liu16TIP} & 4.945 & -0.240 & -53.86 & -59.08 & -63.54 & -66.88 \\
			& & Our & \textbf{2.549} & \textbf{-0.125} & \textbf{-66.48} & \textbf{-70.98} & \textbf{-71.85} & \textbf{-73.85} \\
			
			\hline \multirow{15}{*}{B} & \multirow{3}{*}{\textit{BasketballDrive}} & \cite{Liu16DASC} & 8.923 & -0.244 & -48.53 & -37.13 & -41.05 & -46.89 \\
			& & \cite{Liu16TIP} & 6.018 & -0.141 & -68.50 & -68.55 & -70.30 & -70.70 \\
			& & Our & \textbf{4.265} & \textbf{-0.121} & \textbf{-74.26} & \textbf{-76.18} & \textbf{-76.93} & \textbf{-77.90} \\
			\cline {2-9} & \multirow{3}{*}{\textit{BQTerrace}} & \cite{Liu16DASC} & 6.627 & -0.295 & \textbf{-67.32} & -63.09 & -59.57 & -36.48 \\
			& & \cite{Liu16TIP} & 4.815 & -0.267 & -53.35 & -56.81 & -60.26 & -61.13 \\
			& & Our & \textbf{1.842} & \textbf{-0.086} & -52.02 & \textbf{-65.57} & \textbf{-68.25} & \textbf{-73.05} \\
			\cline {2-9} & \multirow{3}{*}{\textit{Cactus}} & \cite{Liu16DASC} & 7.533 & -0.248 & -38.37 & -40.83 & -43.61 & -51.23 \\
			& & \cite{Liu16TIP} & 6.021 & -0.208 & \textbf{-58.18} & \textbf{-61.01} & -64.94 & -67.78 \\
			& & Our & \textbf{2.265} & \textbf{-0.077} & -48.95 & -53.87 & \textbf{-66.92} & \textbf{-74.09} \\
			\cline {2-9} & \multirow{3}{*}{\textit{Kimono}} & \cite{Liu16DASC} & 5.212 & -0.170 & -34.51 & -43.21 & -49.88 & -63.58 \\
			& & \cite{Liu16TIP} & \textbf{2.382} & \textbf{-0.082} & -70.66 & -72.75 & -73.62 & -73.86 \\
			& & Our & 2.592 & -0.087 & \textbf{-80.40} & \textbf{-84.32} & \textbf{-84.45} & \textbf{-84.97} \\
			\cline {2-9} & \multirow{3}{*}{\textit{ParkScene}} & \cite{Liu16DASC} & 3.630 & -0.149 & -41.69 & -44.79 & -59.98 & -64.92 \\
			& & \cite{Liu16TIP} & 3.417 & -0.135 & -60.27 & -65.10 & -68.57 & -70.16 \\
			& & Our & \textbf{1.956} & \textbf{-0.081} & \textbf{-60.42} & \textbf{-66.15} & \textbf{-70.08} & \textbf{-73.47} \\
			
			\hline \multirow{12}{*}{C} & \multirow{3}{*}{\textit{BasketballDrill}} & \cite{Liu16DASC} & 9.818 & -0.439 & -46.65 & -58.86 & -47.66 & -62.53 \\
			& & \cite{Liu16TIP} & 12.205 & -0.538 & \textbf{-60.29} & \textbf{-62.35} & \textbf{-64.48} & \textbf{-67.20} \\
			& & Our & \textbf{2.863} & \textbf{-0.133} & -42.47 & -45.16 & -62.11 & -62.17 \\
			\cline {2-9} & \multirow{3}{*}{\textit{BQMall}} & \cite{Liu16DASC} & 9.646 & -0.486 & -52.62 & -42.97 & -35.52 & -37.12 \\
			& & \cite{Liu16TIP} & 8.077 & -0.468 & -47.08 & -51.15 & -53.26 & -57.05 \\
			& & Our & \textbf{2.089} & \textbf{-0.111} & \textbf{-52.74} & \textbf{-57.05} & \textbf{-59.39} & \textbf{-64.48} \\
			\cline {2-9} & \multirow{3}{*}{\textit{PartyScene}} & \cite{Liu16DASC} & 7.383 & -0.468 & \textbf{-63.84} & -49.82 & -31.50 & -26.88 \\
			& & \cite{Liu16TIP} & 9.448 & -0.672 & -52.72 & \textbf{-57.51} & \textbf{-59.77} & \textbf{-64.98} \\
			& & Our & \textbf{0.664} & \textbf{-0.044} & -37.27 & -41.94 & -43.78 & -55.00 \\
			\cline {2-9} & \multirow{3}{*}{\textit{RaceHorses}} & \cite{Liu16DASC} & 7.220 & -0.379 & -46.46 & -40.13 & -41.49 & -50.28 \\
			& & \cite{Liu16TIP} & 4.422 & -0.264 & -50.52 & \textbf{-59.30} & \textbf{-59.81} & \textbf{-63.15} \\
			& & Our & \textbf{1.973} & \textbf{-0.108} & \textbf{-54.08} & -54.47 & -58.59 & -61.32 \\
			
			\hline \multirow{12}{*}{D} & \multirow{3}{*}{\textit{BasketballPass}} & \cite{Liu16DASC} & 10.054 & -0.546 & -43.69 & -41.03 & -37.46 & -36.69 \\
			& & \cite{Liu16TIP} & 8.401 & -0.457 & \textbf{-60.24} & \textbf{-62.89} & \textbf{-64.31} & \textbf{-66.67} \\
			& & Our & \textbf{1.842} & \textbf{-0.106} & -54.87 & -56.64 & -56.93 & -57.23 \\
			\cline {2-9} & \multirow{3}{*}{\textit{BlowingBubbles}} & \cite{Liu16DASC} & 6.178 & -0.373 & \textbf{-57.15} & -42.45 & -25.73 & -22.81 \\
			& & \cite{Liu16TIP} & 8.328 & -0.463 & -54.62 & \textbf{-60.45} & \textbf{-62.55} & \textbf{-65.48} \\
			& & Our & \textbf{0.622} & \textbf{-0.039} & -38.08 & -38.66 & -39.52 & -45.91 \\
			\cline {2-9} & \multirow{3}{*}{\textit{BQSquare}} & \cite{Liu16DASC} & 12.342 & -0.876 & \textbf{-61.45} & \textbf{-62.40} & \textbf{-58.99} & -46.86 \\
			& & \cite{Liu16TIP} & 2.563 & -0.211 & -42.55 & -46.05 & -48.37 & \textbf{-49.89} \\
			& & Our & \textbf{0.913} & \textbf{-0.067} & -44.13 & -45.08 & -46.57 & -47.48 \\
			\cline {2-9} & \multirow{3}{*}{\textit{RaceHorses}} & \cite{Liu16DASC} & 8.839 & -0.487 & -43.12 & -40.18 & -39.54 & -38.07 \\
			& & \cite{Liu16TIP} & 4.953 & -0.317 & \textbf{-52.86} & \textbf{-57.32} & \textbf{-58.80} & \textbf{-60.20} \\
			& & Our & \textbf{1.320} & \textbf{-0.075} & -50.91 & -54.17 & -58.23 & -59.72 \\
			
			\hline \multirow{9}{*}{E} & \multirow{3}{*}{\textit{FourPeople}} & \cite{Liu16DASC} & 9.077 & -0.480 & -53.52 & -40.88 & -26.12 & -24.34 \\
			& & \cite{Liu16TIP} & 8.002 & -0.439 & -54.79 & -59.79 & -64.39 & -67.17 \\
			& & Our & \textbf{3.110} & \textbf{-0.171} & \textbf{-65.30} & \textbf{-72.08} & \textbf{-72.99} & \textbf{-74.86} \\
			\cline {2-9} & \multirow{3}{*}{\textit{Johnny}} & \cite{Liu16DASC} & 12.182 & -0.474 & -58.29 & -60.21 & -63.98 & -70.70 \\
			& & \cite{Liu16TIP} & 7.956 & -0.307 & -62.92 & -65.51 & -67.71 & -70.05 \\
			& & Our & \textbf{3.822} & \textbf{-0.153} & \textbf{-70.32} & \textbf{-70.57} & \textbf{-70.73} & \textbf{-71.10} \\
			\cline {2-9} & \multirow{3}{*}{\textit{KristenAndSara}} & \cite{Liu16DASC} & 13.351 & -0.627 & -54.44 & -56.94 & -56.05 & -62.61 \\
			& & \cite{Liu16TIP} & 5.478 & -0.265 & -61.24 & -64.61 & -65.82 & -67.21 \\
			& & Our & \textbf{3.460} & \textbf{-0.169} & \textbf{-72.76} & \textbf{-74.53} & \textbf{-75.98} & \textbf{-76.17} \\

			\hline \multicolumn{2}{|c|}{\multirow{3}{*}{Std. dev.}} & \cite{Liu16DASC} & 2.553 & 0.175 & 9.42 & 10.16 & 13.04 & 15.01 \\
			\multicolumn{2}{|c|}{} & \cite{Liu16TIP} & 2.603 & 0.158 & \textbf{7.10} & \textbf{6.24} & \textbf{6.09} & \textbf{5.63} \\
			\multicolumn{2}{|c|}{} & Our & \textbf{1.020} & \textbf{0.039} & 12.67 & 12.96 & 11.96 & 10.75 \\
			\hline \multicolumn{2}{|c|}{\multirow{3}{*}{Best}} & \cite{Liu16DASC} & 3.630 & -0.149 & -67.32 & -63.09 & -63.98 & -70.70 \\
			\multicolumn{2}{|c|}{} & \cite{Liu16TIP} & 2.382 & -0.082 & -70.66 & -72.75 & -73.62 & -73.86 \\
			\multicolumn{2}{|c|}{} & Our & \textbf{0.622} & \textbf{-0.039} & \textbf{-80.40} & \textbf{-84.32} & \textbf{-84.45} & \textbf{-84.97} \\
			\hline \multicolumn{2}{|c|}{\multirow{3}{*}{Average}} & \cite{Liu16DASC} & 8.559 & -0.419 & -50.05 & -46.72 & -43.09 & -45.01 \\
			\multicolumn{2}{|c|}{} & \cite{Liu16TIP} & 6.189 & -0.316 & -56.41 & -60.23 & -62.62 & -65.04 \\
			\multicolumn{2}{|c|}{} & Our & \textbf{2.247} & \textbf{-0.104} & \textbf{-56.92} & \textbf{-60.38} & \textbf{-63.61} & \textbf{-66.47} \\
			\hline
			
		\end{tabular}
	\end{center}
\end{table}

\begin{figure*}
	\centering
	\begin{minipage}{0.32\linewidth}
		\centering
		\includegraphics[width=1.05\linewidth]{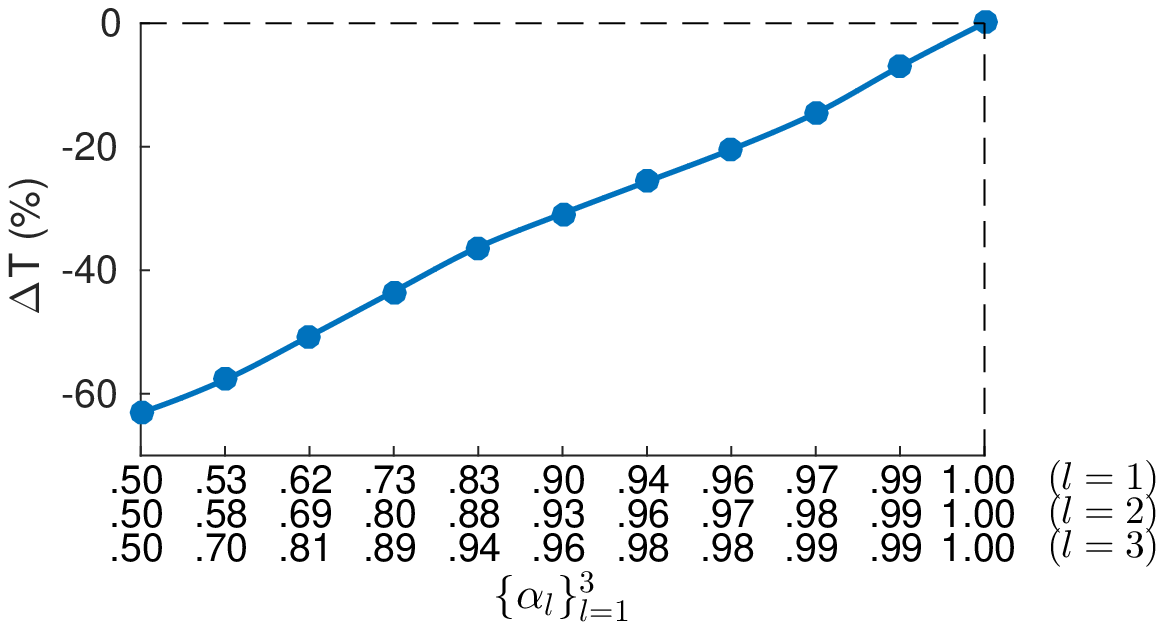}\\
		%\centerline{\footnotesize{(a)}}\medskip
	\end{minipage}
	\hfill
	\begin{minipage}{0.32\linewidth}
		\centering
		\includegraphics[width=1.05\linewidth]{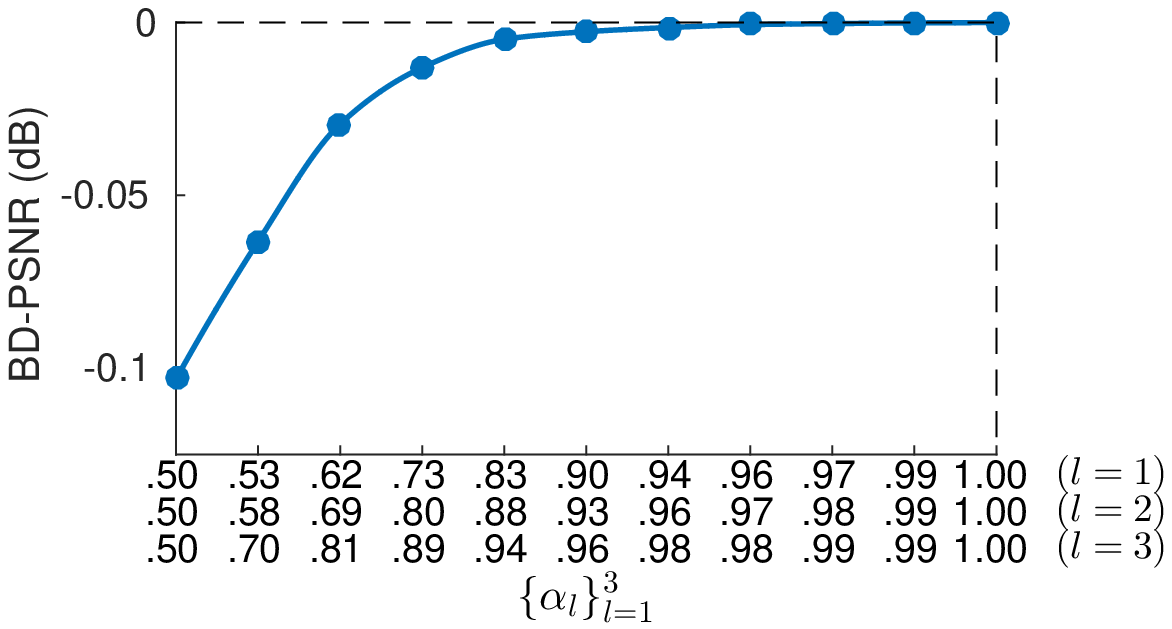}\\
		%\centerline{\footnotesize{(b)}}\medskip
	\end{minipage}
	\hfill
	\begin{minipage}{0.32\linewidth}
		\centering
		\includegraphics[width=1.05\linewidth]{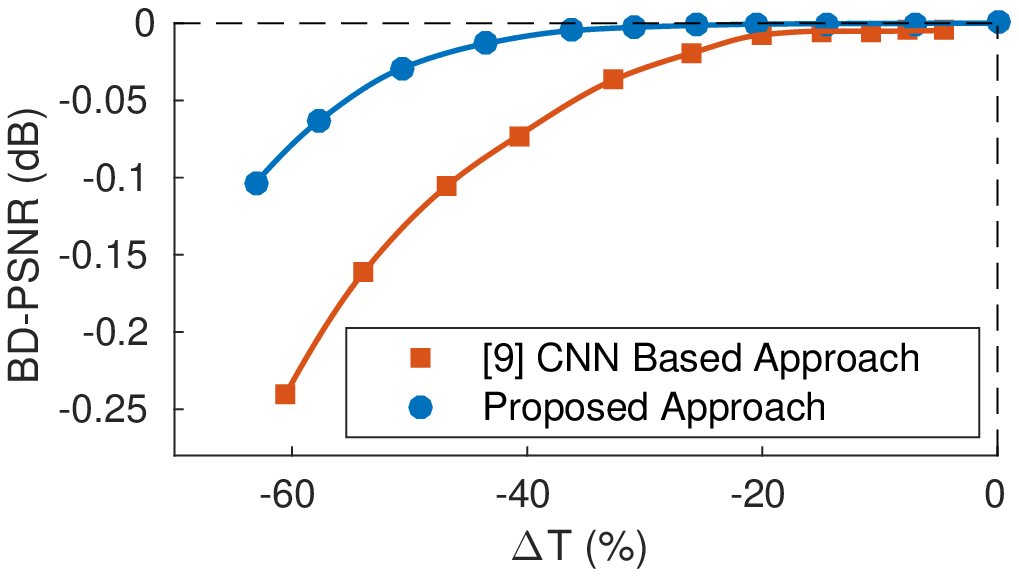}\\
		%\centerline{\footnotesize{(c)}}\medskip
	\end{minipage}
	\caption{\footnotesize CRD performance for intra-mode HEVC complexity reduction. Left and middle: Complexity reduction and RD performance at different uncertain zones. No that the labels on horizontal axes are upper thresholds $\{\alpha_l\}_{l=1}^3$, and the corresponding lower thresholds are $\bar{\alpha}_l=1-\alpha_l$ for $l\in\{1,2,3\}$. Right: Comparison of CRD performance between both approaches.}\label{fig:tradeoff-intra}
\end{figure*}

\textbf{Evaluation on training performance.} When predicting the CU partition of intra-mode HEVC, our approach relies on the training model of ETH-CNN, which is optimized by \eqref{eq:loss-sum-intra} over the training set. Therefore, it is necessary to evaluate the training performance of our approach at intra-mode. Fig. \ref{fig:train-loss-accu}-(a) shows the training loss along with iterations during the model training of ETH-CNN. Note that the training loss is obtained through calculating \eqref{eq:loss-sum-intra} at each training iteration. We can see from this figure that the loss converges after $5\times 10^5$ iterations, when training the ETH-CNN model at the AI configuration.

\textbf{Evaluation on prediction accuracy.}
First, we evaluate the accuracy of the CU partition predicted by our ETH-CNN approach, when compressing images/sequences with intra-mode HEVC.
In our experiments, the accuracy is averaged over the CU partition results of 200 test images and 18 test sequences, which are compressed by HM 16.5 at the AI configuration and four QP values $\{22,27,32,37\}$. We find from the experimental results that the average accuracy rates for predicting the intra-mode CU partition at levels 1, 2 and 3 are $90.98\%$, $86.42\%$ and $80.42\%$, respectively. These accuracy rates are rather high, thus leading to a slight loss of RD performance, as discussed below.

\textbf{Comparison of complexity reduction.}
Next, we compare our approach with \cite{Liu16DASC} and \cite{Liu16TIP} in complexity reduction, when compressing images/sequences by HEVC at intra-mode. Tables \ref{tab:result-image-intra} and \ref{tab:result-video-intra} tabulate the complexity reduction results in $\Delta T$ for 200 test images and 18 test sequences, respectively.
As observed in Table \ref{tab:result-image-intra}, at QP $= 22, 27, 32$ and $37$, our approach reduces the encoding complexity by $64.01\%$, $66.09\%$, $68.11\%$ and $70.52\%$ on average, outperforming the $58.36\%, 56.27\%, 54.23\%$ and $55.69\%$ complexity reductions in \cite{Liu16DASC} and the $ 58.05\%, 61.66\%, 65.25\%$ and $67.26\%$ complexity reductions in \cite{Liu16TIP}.
In addition, for most scenarios with different QPs and sequences, our approach achieves a better complexity reduction than the other two approaches.
Note that our ETH-CNN based approach requires less time than the shallow CNN approach \cite{Liu16TIP}, because \cite{Liu16TIP} requires an RDO search for the decision on splitting from $64\times64$ to $32\times 32$.
Our approach outperforms \cite{Liu16DASC} and \cite{Liu16TIP} in terms of maximal $\Delta{}T$, for compressing the test images and video sequences with intra-mode HEVC. More importantly, when applying our approach and \cite{Liu16TIP} in intra-mode compression of the test images and video sequences, the standard deviations of $\Delta{}T$ are less than $\frac{1}{5}\cdot|\Delta{}T|$, indicating the stability of both approaches in complexity reduction.
For video sequences, similar results can be found in Table \ref{tab:result-video-intra}.
From Tables \ref{tab:result-image-intra} and \ref{tab:result-video-intra}, we can further observe that the gap in the time reduction between our and other approaches increases along with increasing QP. This is most likely because the increased QP leads to more large CUs, such that early termination in our ETH-CNN model works for more CTUs, resulting in decreased running time.
In summary, our approach is capable of improving the time efficiency of intra-mode HEVC.
%Since ETH-CNN is capable to predict the CU partition accurately, here the bi-threshold decision mentioned in Section \ref{sec:bi-threshold} is not adopted. In other words, $\{\bar{\alpha}_{l}\}_{l=1}^3$ and $\{\alpha_{l}\}_{l=1}^3$ are all 0.5, the default value, for achieving the best complexity reduction performance.

%Our approach is also able to reduce more time averagely, on encoding all images of the test set from our CPH-Intra database, compared with \cite{Zhang15TIP} and \cite{Liu16TIP}.
% ------------------
%We can further find that the gap of time saving between our and other two approaches becomes larger when QP increases. This is because only one or two deep CNN models are applied in our three-level classifier (corresponding to classifier $S_1$ or $S_2$) for more CTUs at decreasing bit-rate, which leads to more large CUs.  % ICME version
%In both tables above, we can further find that the gap of time saving between our and other two approaches becomes larger when QP increases. This is because early termination in our ETH-CNN model works for more CTUs at decreasing bit-rate, which leads to more large CUs.  % current version
%% ------------------
%However, for low resolution images encoded at a high bit-rate, our approach may have a higher computational complexity than \cite{Zhang15TIP} or \cite{Liu16TIP}. The reason is that CTUs prefer to have small CUs, when resolution is low and bit-rate is high.
%In a word, our approach is capable of improving the time efficiency of HEVC intra-mode coding.

\textbf{Comparison of RD performance.}
The reduced complexity of intra-mode HEVC is at the expense of an RD performance loss. Here, we compare the RD performance of our and the other two approaches in terms of BD-BR and BD-PSNR. Tables \ref{tab:result-image-intra} and \ref{tab:result-video-intra} report the BD-BR and BD-PSNR under the three approaches, with the original HM as the anchor. We can see from these tables that the BD-BR increase in our ETH-CNN approach is on average $1.386\%$ for images and $2.247\%$ for sequences, which significantly outperforms \cite{Liu16DASC} ($4.945\%$ for images and $8.559\%$ for sequences) and \cite{Liu16TIP} ($2.353\%$ for images and $6.189\%$ for sequences). In addition, our approach incurs $-0.081$ dB and $-0.104$ dB BD-PSNR for images and sequences, respectively, which are far better than the $-0.284$ dB and $-0.419$ dB of \cite{Liu16DASC}, and the $-0.155$ dB and $-0.316$ dB of \cite{Liu16TIP}.
Our approach is also superior to \cite{Liu16DASC} and \cite{Liu16TIP}, in terms of both standard deviations and best values of BD-BR and BD-PSNR.
Thus, our approach performs best among the three approaches in terms of RD performance. The RD improvement of our approach is mainly due to the high prediction accuracy of the CU partition, benefiting from the deep ETH-CNN structure with sufficient parameters learned from our large-scale CPH-Intra database.

\textbf{Comparison of complexity-RD performance.}
As mentioned in Section \ref{sec:bi-threshold}, the bi-threshold scheme can control the RD performance and encoding complexity. Therefore, we changed the uncertain zone $[\bar{\alpha}_{l}, \alpha_{l}]$ of the bi-threshold scheme to assess the complexity-RD (CRD) performance of our approach.
As mentioned in Section \ref{sec:bi-threshold}, $[\bar{\alpha}_1, \alpha_1]\subset[\bar{\alpha}_2, \alpha_2]\subset[\bar{\alpha}_3, \alpha_3]$ and $\bar{\alpha}_l=1-\alpha_l$ exist for our bi-threshold scheme. Therefore, the specific values of $\{[\bar{\alpha}_l, \alpha_l]\}_{l=1}^3$ at each level $l$ are determined by power functions for simplicity:
\begin{equation}\label{eq:bi-threshold}
	\left\{
	\begin{array}{lcl}
		\alpha_l=0.5+0.5d^{2-0.5l}\\
		\bar\alpha_l=0.5-0.5d^{2-0.5l}\\
	\end{array}\quad  l\in\{1,2,3\}, \right.
\end{equation}
where $d\in[0,1]$ is the width of the uncertain zone at level $l=2$. Specifically, Fig. \ref{fig:tradeoff-intra}-left and -middle show the results of the encoding complexity reduction and RD performance degradation at different uncertain zones by adjusting $d$, for intra-mode HEVC.
The results are averaged over the JCT-VC test sequences at QP $= 22, 27, 32$, and $37$. In this figure, the encoding complexity reduction is measured by $\Delta T$ (i.e., encoding time saving) and the RD performance degradation is assessed by BD-PSNR.
We can see from Fig. \ref{fig:tradeoff-intra}-left and -middle that the absolute values of both $\Delta T$ and BD-PSNR are increased when the uncertain zone of $[\bar{\alpha_l},\alpha_l]$ becomes narrower. This indicates that the trade-off between the RD performance and encoding complexity can be controlled by changing the uncertain zone $[\bar{\alpha_l},\alpha_l]$. In Fig. \ref{fig:tradeoff-intra}-right, we also plot the CRD curves of our approach and the state-of-the-art CNN based approach \cite{Liu16TIP} by plotting the values of $\Delta T$ and BD-PSNR, which are obtained at different uncertain zones of $[\bar{\alpha}_{l}, \alpha_{l}]$. This figure shows that the RD performance is improved at less reduction of encoding complexity. This figure further shows that our approach achieves a better RD performance than \cite{Liu16TIP} at the same encoding complexity reduction.

\subsection{Performance Evaluation at Inter-mode}
\label{sec:PEiner}

\begin{table}
	\tiny
	\newcommand{\tabincell}[2]{\begin{tabular}{@{}#1@{}}#2\end{tabular}}
	\begin{center}
		\caption{\footnotesize{Results for sequences of the JCT-VT test set (LDP)}}
		\label{tab:result-video-inter}
		\begin{tabular}{|c|c|c|c|c|c|c|c|c|}
			\hline \multirow{2}{*}{Class} & \multirow{2}{*}{Sequence} & \multirow{2}{*}{Appr.} & \multirow{2}{*}{\tabincell{c}{\hspace{-0.5em}BD-BR\hspace{-0.5em} \\(\%)}} & \multirow{2}{*}{\tabincell{c}{\hspace{-0.5em}BD-PSNR\hspace{-0.5em} \\(dB)}} & \multicolumn{4}{|c|}{$\Delta{}T$ (\%)} \\
			\cline{6-9} & & & & & QP=22 & QP=27 & QP=32 & QP=37 \\
			
			\hline \multirow{8}{*}{A} & \multirow{4}{*}{\textit{PeopleOnStreet}} & \cite{Zhang15TIP} & 1.456 & -0.062 & -35.26 & -42.53 & -45.40 & -46.32 \\
			& & \cite{Correa15TCSVT} & 9.813 & -0.400 & -29.05 & -26.39 & -31.28 & -39.18 \\
			& & \cite{Mallikarachchi16TCSVT} & 4.292 & -0.180 & -39.32 & -40.46 & -41.85 & -43.99 \\
			& & Our & \textbf{1.052} & \textbf{-0.045} & \textbf{-40.19} & \textbf{-45.73} & \textbf{-50.02} & \textbf{-52.07} \\
			\cline {2-9} & \multirow{4}{*}{\textit{Traffic}} & \cite{Zhang15TIP} & 2.077 & -0.056 & -43.66 & -57.25 & -62.91 & -58.04 \\
			& & \cite{Correa15TCSVT} & 4.639 & -0.119 & -38.30 & -50.70 & -61.39 & -68.60 \\
			& & \cite{Mallikarachchi16TCSVT} & 6.983 & -0.178 & \textbf{-48.71} & -52.22 & -56.96 & -60.03 \\
			& & Our & \textbf{1.990} & \textbf{-0.052} & -47.43 & \textbf{-59.39} & \textbf{-65.00} & \textbf{-70.59} \\
			
			\hline \multirow{20}{*}{B} & \multirow{4}{*}{\textit{BasketballDrive}} & \cite{Zhang15TIP} & \textbf{2.246} & -0.055 & -33.61 & -48.72 & -50.70 & -47.62 \\
			& & \cite{Correa15TCSVT} & 6.372 & -0.145 & -33.90 & -45.16 & -52.16 & -58.47 \\
			& & \cite{Mallikarachchi16TCSVT} & 3.991 & -0.092 & -40.77 & -44.36 & -46.67 & -48.67 \\
			& & Our & 2.268 & \textbf{-0.052} & \textbf{-41.16} & \textbf{-55.34} & \textbf{-60.18} & \textbf{-66.69} \\
			\cline {2-9} & \multirow{4}{*}{\textit{BQTerrace}} & \cite{Zhang15TIP} & 1.733 & -0.029 & -32.15 & -41.84 & -51.59 & -49.80 \\
			& & \cite{Correa15TCSVT} & 1.934 & -0.028 & -15.32 & -35.77 & -58.41 & -71.95 \\
			& & \cite{Mallikarachchi16TCSVT} & 2.892 & -0.042 & \textbf{-45.49} & -47.80 & -53.76 & -56.01 \\
			& & Our & \textbf{1.090} & \textbf{-0.017} & -44.39 & \textbf{-55.72} & \textbf{-67.97} & \textbf{-71.97} \\
			\cline {2-9} & \multirow{4}{*}{\textit{Cactus}} & \cite{Zhang15TIP} & \textbf{2.005} & -0.044 & -33.24 & -47.21 & -52.77 & -50.00 \\
			& & \cite{Correa15TCSVT} & 5.125 & -0.105 & -25.09 & -42.48 & -53.95 & -62.03 \\
			& & \cite{Mallikarachchi16TCSVT} & 5.516 & -0.111 & \textbf{-43.92} & -49.32 & -52.44 & -53.60 \\
			& & Our & 2.071 & \textbf{-0.043} & -41.51 & \textbf{-55.93} & \textbf{-61.56} & \textbf{-68.51} \\
			\cline {2-9} & \multirow{4}{*}{\textit{Kimono}} & \cite{Zhang15TIP} & 2.219 & -0.066 & -41.68 & -52.39 & -53.98 & -53.27 \\
			& & \cite{Correa15TCSVT} & 4.382 & -0.128 & -43.06 & -51.24 & -57.77 & \textbf{-66.33} \\
			& & \cite{Mallikarachchi16TCSVT} & 3.239 & -0.097 & \textbf{-46.21} & -47.70 & -49.76 & -50.01 \\
			& & Our & \textbf{1.497} & \textbf{-0.048} & -45.58 & \textbf{-54.99} & \textbf{-59.53} & -64.02 \\
			\cline {2-9} & \multirow{4}{*}{\textit{ParkScene}} & \cite{Zhang15TIP} & 1.559 & -0.044 & -38.88 & -47.36 & -54.17 & -49.50 \\
			& & \cite{Correa15TCSVT} & 3.493 & -0.098 & -29.65 & -39.63 & -53.37 & -64.09 \\
			& & \cite{Mallikarachchi16TCSVT} & 4.493 & -0.126 & -43.93 & -45.53 & -49.39 & -51.86 \\
			& & Our & \textbf{1.474} & \textbf{-0.042} & \textbf{-46.48} & \textbf{-56.08} & \textbf{-63.25} & \textbf{-69.10} \\
			
			\hline \multirow{16}{*}{C} & \multirow{4}{*}{\textit{BasketballDrill}} & \cite{Zhang15TIP} & \textbf{1.599} & \textbf{-0.058} & -36.96 & -43.63 & -45.48 & -42.93 \\
			& & \cite{Correa15TCSVT} & 7.485 & -0.266 & -29.66 & -34.97 & -43.46 & -53.26 \\
			& & \cite{Mallikarachchi16TCSVT} & 4.055 & -0.147 & -44.00 & -41.93 & -41.56 & -41.99\\
			& & Our & 1.953 & -0.072 & \textbf{-49.75} & \textbf{-52.75} & \textbf{-57.63} & \textbf{-60.66} \\
			\cline {2-9} & \multirow{4}{*}{\textit{BQMall}} & \cite{Zhang15TIP} & \textbf{1.604} & \textbf{-0.058} & -39.48 & -43.32 & -47.54 & -43.84 \\
			& & \cite{Correa15TCSVT} & 6.801 & -0.239 & -34.13 & -36.04 & -46.16 & -55.83 \\
			& & \cite{Mallikarachchi16TCSVT} & 4.092 & -0.147 & -40.73 & -39.05 & -39.04 & -39.92\\
			& & Our & 1.914 & -0.071 & \textbf{-42.02} & \textbf{-48.58} & \textbf{-54.50} & \textbf{-57.86} \\
			\cline {2-9} & \multirow{4}{*}{\textit{PartyScene}} & \cite{Zhang15TIP} & 1.301 & -0.047 & -31.34 & -34.49 & -36.71 & -33.40 \\
			& & \cite{Correa15TCSVT} & 2.988 & -0.107 & -12.64 & -18.14 & -28.75 & -43.19 \\
			& & \cite{Mallikarachchi16TCSVT} & 2.469 & -0.094 & \textbf{-42.75} & -38.35 & -37.40 & -37.89 \\
			& & Our & \textbf{1.011} & \textbf{-0.039} & -38.31 & \textbf{-41.77} & \textbf{-49.00} & \textbf{-58.24} \\
			\cline {2-9} & \multirow{4}{*}{\textit{RaceHorses}} & \cite{Zhang15TIP}  & 2.085 & -0.072 & -36.39 & -39.58 & -39.18 & -37.80 \\
			& & \cite{Correa15TCSVT} & 6.204 & -0.215 & -20.17 & -25.83 & -33.28 & -40.32 \\
			& & \cite{Mallikarachchi16TCSVT} & 3.016 & -0.105 & -36.14 & -33.56 & -31.90 & -32.97 \\
			& & Our & \textbf{0.872} & \textbf{-0.032} & \textbf{-39.65} & \textbf{-43.25} & \textbf{-48.65} & \textbf{-53.34} \\
			
			\hline \multirow{16}{*}{D} & \multirow{4}{*}{\textit{BasketballPass}} & \cite{Zhang15TIP} & \textbf{1.321} & \textbf{-0.059} & -31.74 & -33.89 & -34.58 & -33.59 \\
			& & \cite{Correa15TCSVT} & 6.995 & -0.310 & -31.66 & -29.44 & -33.39 & -41.93 \\
			& & \cite{Mallikarachchi16TCSVT}  & 1.949 & -0.086 & -36.80 & -31.04 & -29.01 & -25.80 \\
			& & Our & 1.453 & -0.066 & \textbf{-43.69} & \textbf{-46.85} & \textbf{-50.66} & \textbf{-54.97} \\
			\cline {2-9} & \multirow{4}{*}{\textit{BlowingBubbles}} & \cite{Zhang15TIP} & \textbf{0.974} & \textbf{-0.034} & -23.81 & -27.92 & -26.40 & -18.11 \\
			& & \cite{Correa15TCSVT} & 2.180 & -0.078 & -14.18 & -19.91 & -31.71 & -40.22 \\
			& & \cite{Mallikarachchi16TCSVT} & 2.479 & -0.086 & \textbf{-35.52} & -31.21 & -29.04 & -26.03 \\
			& & Our & 1.292 & -0.044 & -33.63 & \textbf{-41.40} & \textbf{-50.21} & \textbf{-57.26} \\
			\cline {2-9} & \multirow{4}{*}{\textit{BQSquare}} & \cite{Zhang15TIP} & 1.175 & -0.036 & -24.80 & -31.02 & -31.84 & -27.88 \\
			& & \cite{Correa15TCSVT} & 2.090 & -0.062 & -13.96 & -20.62 & -43.86 & -45.72 \\
			& & \cite{Mallikarachchi16TCSVT} & 1.998 & -0.067 & \textbf{-45.19} & -34.69 & -26.46 & -26.09 \\
			& & Our & \textbf{0.770} & \textbf{-0.028} & -40.24 & \textbf{-40.61} & \textbf{-49.33} & \textbf{-57.47} \\
			\cline {2-9} & \multirow{4}{*}{\textit{RaceHorses}} & \cite{Zhang15TIP} & 1.306 & -0.054 & -29.48 & -29.34 & -28.37 & -32.78 \\
			& & \cite{Correa15TCSVT} & 6.867 & -0.281 & -19.58 & -20.26 & -25.37 & -33.07 \\
			& & \cite{Mallikarachchi16TCSVT} & 2.240 & -0.095 & -34.81 & -27.72 & -22.92 & -20.09 \\
			& & Our & \textbf{1.112} & \textbf{-0.047} & \textbf{-37.07} & \textbf{-39.26} & \textbf{-43.14} & \textbf{-47.99} \\
			
			\hline \multirow{12}{*}{E} & \multirow{4}{*}{\textit{FourPeople}} & \cite{Zhang15TIP} & 2.593 & -0.082 & -58.11 & -67.79 & -71.85 & -67.62 \\
			& & \cite{Correa15TCSVT} & 5.492 & -0.165 & \textbf{-59.74} & \textbf{-69.08} & \textbf{-74.29} & \textbf{-79.19} \\
			& & \cite{Mallikarachchi16TCSVT} & 3.974 & -0.122 & -53.81 & -56.60 & -57.11 & -58.12 \\
			& & Our & \textbf{1.837} & \textbf{-0.052} & -53.36 & -64.29 & -67.89 & -71.97 \\
			\cline {2-9} & \multirow{4}{*}{\textit{Johnny}} & \cite{Zhang15TIP} & 3.024 & -0.058 & -59.16 & -72.00 & -74.29 & -73.91 \\
			& & \cite{Correa15TCSVT} & 3.769 & -0.079 & \textbf{-65.45} & \textbf{-75.72} & \textbf{-81.45} & \textbf{-84.83} \\
			& & \cite{Mallikarachchi16TCSVT} & 3.452 & -0.065 & -55.45 & -58.49 & -59.50 & -59.82 \\
			& & Our & \textbf{1.691} & \textbf{-0.038} & -51.99 & -67.68 & -71.34 & -74.97 \\
			\cline {2-9} & \multirow{4}{*}{\textit{KristenAndSara}} & \cite{Zhang15TIP} & 2.098 & -0.055 & -59.90 & -69.46 & -73.25 & -70.55 \\
			& & \cite{Correa15TCSVT} & 4.293 & -0.113 & \textbf{-63.34} & \textbf{-71.53} & \textbf{-77.57} & \textbf{-81.40} \\
			& & \cite{Mallikarachchi16TCSVT} & 3.964 & -0.104 & -52.68 & -55.75 & -58.98 & -58.77 \\
			& & Our & \textbf{1.558} & \textbf{-0.045} & -52.69 & -68.77 & -72.19 & -75.27 \\
			
			\hline \multicolumn{2}{|c|}{\multirow{4}{*}{Std. dev.}} & \cite{Zhang15TIP} & 0.534  &  \textbf{0.013}  & 10.84  & 13.44 &  14.82  & 14.88 \\
			\multicolumn{2}{|c|}{} & \cite{Correa15TCSVT} & 2.146  &  0.100  & 16.62  & 18.13 &  17.18  & 16.08 \\
			\multicolumn{2}{|c|}{} & \cite{Mallikarachchi16TCSVT} & 1.282  &  0.037  &  6.17  &  9.34 &  12.04  & 13.30 \\
			\multicolumn{2}{|c|}{} & Our & \textbf{0.448}  &  0.014  &  \textbf{5.60}  &  \textbf{9.27} &   \textbf{8.84}  &  \textbf{8.45} \\
			
			\hline \multicolumn{2}{|c|}{\multirow{4}{*}{Best}} & \cite{Zhang15TIP} & 0.974  & -0.029 & -59.90 & -72.00 & -74.29 & -73.91 \\		\multicolumn{2}{|c|}{} & \cite{Correa15TCSVT} & 1.934  & -0.028 & \textbf{-65.45} & \textbf{-75.72} & \textbf{-81.45} & \textbf{-84.83} \\
			\multicolumn{2}{|c|}{} & \cite{Mallikarachchi16TCSVT} & 1.949  & -0.042 & -55.45 & -58.49 & -59.50 & -60.03 \\
			\multicolumn{2}{|c|}{} & Our & \textbf{0.770}  & \textbf{-0.017} & -53.36 & -68.77 & -72.19 & -75.27 \\
			
			\hline \multicolumn{2}{|c|}{\multirow{4}{*}{Average}} & \cite{Zhang15TIP} & 1.799 & -0.054 & -38.31 & -46.10 & -48.95 & -46.50 \\
			\multicolumn{2}{|c|}{} & \cite{Correa15TCSVT} & 5.051 & -0.163 & -32.16 & -39.59 & -49.31 & -57.20 \\
			\multicolumn{2}{|c|}{} & \cite{Mallikarachchi16TCSVT} & 3.616 & -0.108 & -43.67 & -43.10 & -43.54 & -43.98 \\
			\multicolumn{2}{|c|}{} & Our & \textbf{1.495} & \textbf{-0.046} & \textbf{-43.84} & \textbf{-52.13} & \textbf{-57.89} & \textbf{-62.94} \\
			\hline
			
		\end{tabular}
	\end{center}
\end{table}

\textbf{Evaluation on training performance.} First, we evaluate the training loss at different iterations when training the ETH-CNN and ETH-LSTM models for inter-mode HEVC complexity reduction. Fig. \ref{fig:train-loss-accu}-(a) and -(b) illustrate the training loss curves for ETH-CNN and ETH-LSTM, respectively. Here, the loss values are obtained from \eqref{eq:loss-sum-intra} for training the ETH-CNN model and from \eqref{eq:loss-sum-inter} for training the ETH-LSTM model.  We can see that both the ETH-CNN and LSTM models converge at a fast speed. Additionally, at the LDP configuration, the training loss of the ETH-LSTM model converges to a smaller value in predicting the CU partition of HEVC, compared with the ETH-CNN model.

\textbf{Evaluation on prediction accuracy.}
First, we evaluate the accuracy of the CU partition predicted by ETH-LSTM when compressing 18 test sequences using HEVC at the LDP configuration.
The experimental results show that the average accuracy rates of the inter-mode CU partition at three levels are $93.89\%$, $88.01\%$ and $80.91\%$. These values are much higher than the $90.21\%$, $82.62\%$ and $79.42\%$ of the three-level joint SVM classifier reported in \cite{Zhang15TIP}. Moreover, the accuracy rates of the inter-mode CU partition are higher than those of the intra-mode CU partition in our deep learning approach (reported in Section \ref{sec:PEintra}) because of the temporal correlation of the CU partition learned in ETH-LSTM.

\textbf{Evaluation on complexity reduction.}
Next, we compare our approach with the latest work of \cite{Zhang15TIP}, \cite{Correa15TCSVT} and \cite{Mallikarachchi16TCSVT} for encoding complexity reduction in inter-mode HEVC. In Table \ref{tab:result-video-inter}, $\Delta T$ illustrates the complexity reduction rates of 18 test sequences.
As seen in this table, at QP $= 22, 27, 32$ and $37$, our approach can reduce the encoding complexity by $43.84\%$, $52.13\%$, $57.89\%$ and $62.94\%$ on average. Such complexity reduction rates are superior to the $38.31\%, 46.10\%, 48.95\%$ and $46.50\%$ of \cite{Zhang15TIP}, the $32.16\%, 39.59\%, 49.31\%$ and $57.20\%$ of \cite{Correa15TCSVT}, and the $43.67\%, 43.10\%, 43.54\%$ and $43.98\%$ of \cite{Mallikarachchi16TCSVT}.
In addition, our approach is able to reduce more complexity than the other three approaches on most sequences under different QPs and resolutions.
This is mainly because our approach is able to predict all the CU partition of an entire CTU at a time in the form of HCPM, whereas the other approaches need to run the classifier several times or to check the RDO of some CU partition in splitting a CTU.
We can further see from Table \ref{tab:result-video-inter} that our approach at the LDP configuration has the smallest standard deviations of $\Delta{}T$ among all four approaches.
This implies that our approach is robust in reducing complexity of inter-mode HEVC. However, the maximal complexity reduction of our approach is less than that of \cite{Correa15TCSVT} for inter-mode HEVC. Again, this implies the robustness of our approach in complexity reduction of inter-mode HEVC, since our approach has more average complexity reduction than \cite{Correa15TCSVT}.

\textbf{Evaluation on RD performance.}
In addition to complexity reduction, RD performance is also a critical metric for evaluation. In our experiments, we compare RD performance of all four approaches, in terms of BD-BR and BD-PSNR. Table \ref{tab:result-video-inter} tabulates BD-BR and BD-PSNR results of four approaches, with the original HM as anchor. The BD-BR of our ETH-LSTM approach is averagely $1.495\%$, better than $1.799\%$ of \cite{Zhang15TIP}, $5.051\%$ of \cite{Correa15TCSVT} and $3.616\%$ of \cite{Mallikarachchi16TCSVT}.
On the other hand, the BD-PSNR of our approach is $-0.046$ dB, which is superior to $-0.054$dB of \cite{Zhang15TIP}, $-0.163$dB of \cite{Correa15TCSVT} and $-0.108$dB of \cite{Mallikarachchi16TCSVT}.
Also, Table \ref{tab:result-video-inter} shows that our approach is generally superior to \cite{Zhang15TIP}, \cite{Correa15TCSVT} and \cite{Mallikarachchi16TCSVT} for both standard deviations and best values of the RD loss (evaluated by BD-BR and BD-PSNR).
Therefore, our approach performs best in terms of RD, among all four approaches. It is mainly attributed to the high accuracy of the CU partition predicted by ETH-LSTM as discussed above.

\subsection{Analysis on Running Time}
\label{sec:runningtime}

\begin{figure}
	\centering
	\begin{minipage}{0.56\linewidth}
		\centering
		\includegraphics[width=1\linewidth]{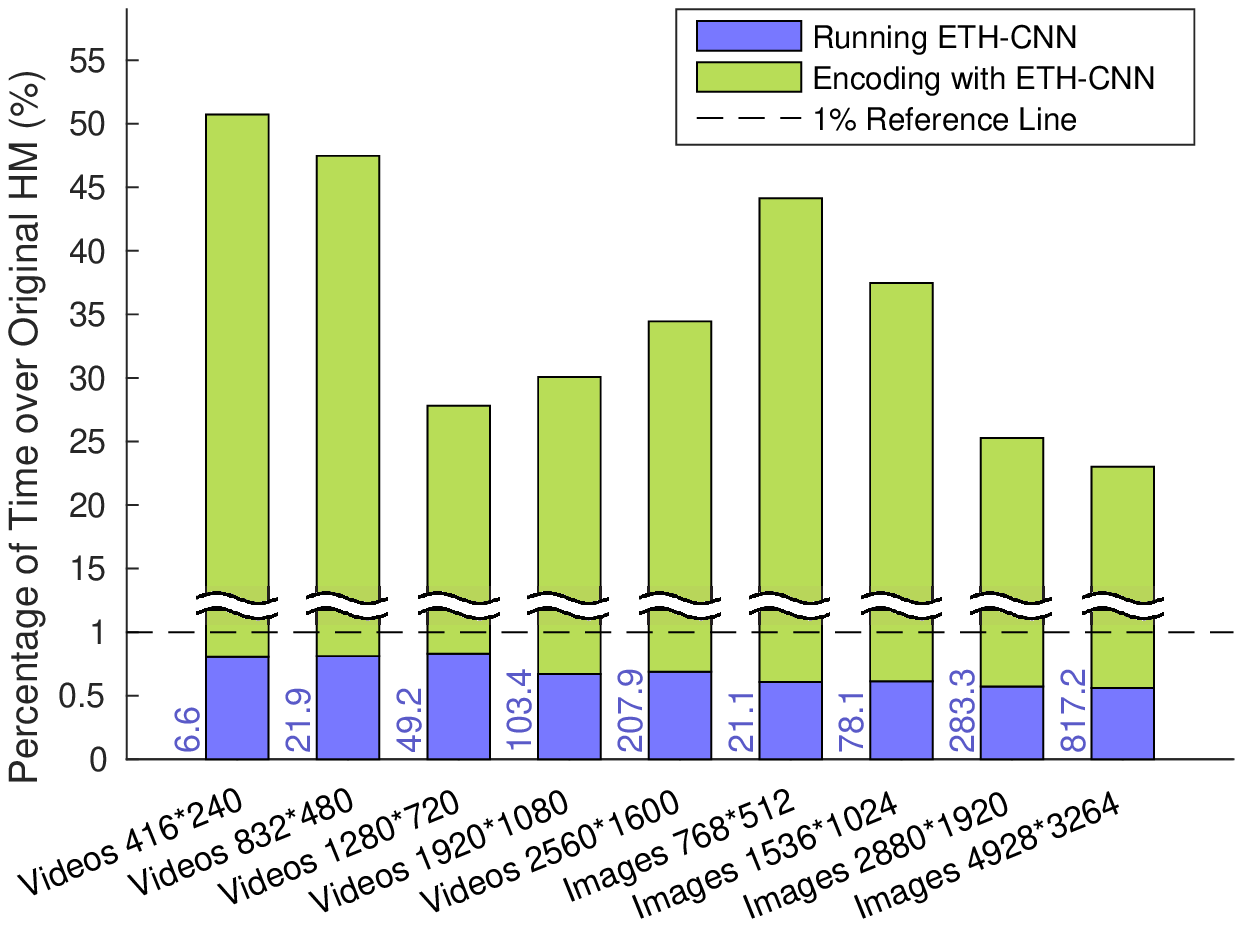}\\
		\centerline{\footnotesize{(a) AI configuration.}}\medskip
	\end{minipage}
	\hfill
	\begin{minipage}{0.395\linewidth}
		\centering
		\includegraphics[width=1\linewidth]{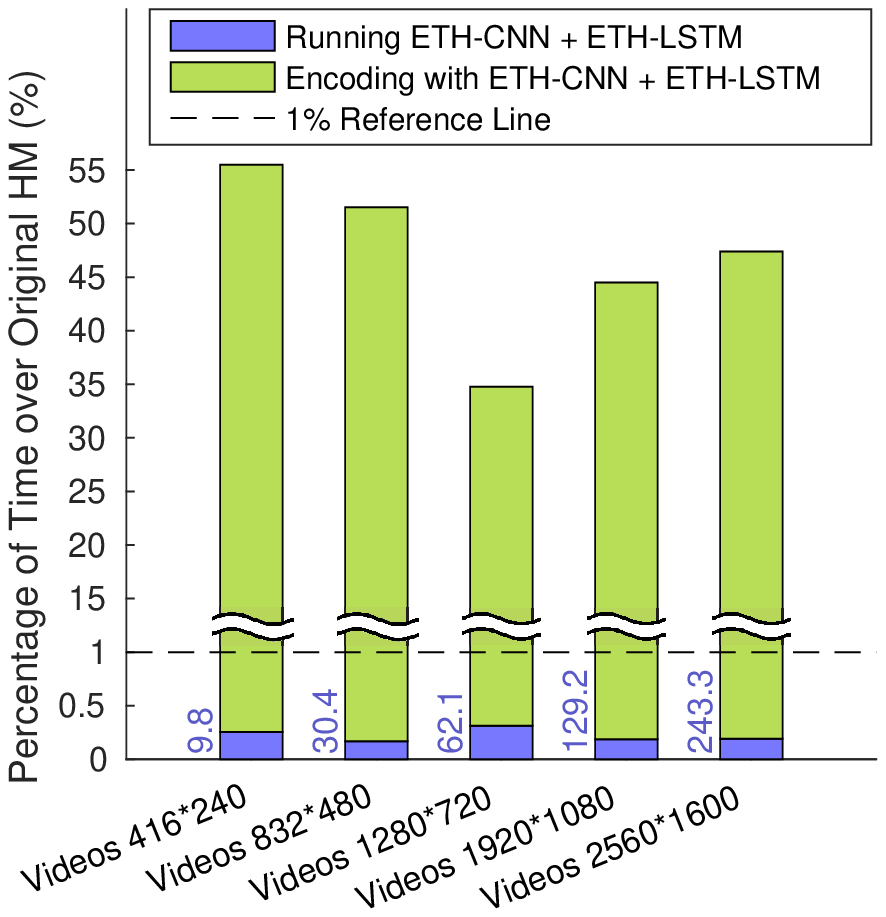}\\
		\centerline{\footnotesize{(b) LDP configuration.}}\medskip
	\end{minipage}
	\caption{\footnotesize{Time percentages for running our deep learning models and encoding videos/images by HM with our complexity reduction approach, with the original HM 16.5 as an anchor. Note the number in blue font next to each bar indicates the real running time of ETH-CNN and ETH-CNN + ETH-LSTM (in millisecond) for one frame.}}\label{fig:time-intra}
\end{figure}

We analyze the running time of our deep learning approach by comparing with that of the original HM 16.5 \cite{HM}.
Fig. \ref{fig:time-intra} shows the time percentages of running ETH-CNN and ETH-CNN + ETH-LSTM relative to the original HM. 
It also plots the time percentages of encoding with ETH-CNN and ETH-CNN + ETH-LSTM, with the original HM as an anchor. 
The number in blue font next to each bar in Fig.10 indicates the absolute time\footnote{Note that only one thread of CPU is used for the above running time, in accord with the HM platform that is also implemented in one thread. If enabling the multi-thread parallelism of CPU, the running time can be drastically reduced.} of running ETH-CNN and ETH-CNN + ETH-LSTM (in millisecond) for one frame.
The results of Fig. \ref{fig:time-intra} are obtained by averaging over all test images/sequences with the same resolution at four QP values $\{22,27,32,37\}$. We can see from Fig. \ref{fig:time-intra}-(a) that ETH-CNN consumes less than $1\%$ of the time required by the original HM. Moreover, ETH-CNN takes up a small portion of time when compressing images/sequences by HM with our complexity reduction approach. Hence, ETH-CNN\footnote{Although running ETH-CNN requires little time for each video frame, a huge amount of time is taken for training the ETH-CNN model, as it is trained from large-scale data. This can be done off-line, making our ETH-CNN practical for complexity reduction in HEVC.} introduces little time overhead in reducing the complexity of intra-mode HEVC.

%Fig. \ref{fig:time-intra}-(b) illustrates the running time percentages of ETH-CNN and ETH-LSTM compared to the original HM, for inter-mode HEVC complexity reduction. Additionally, Fig. \ref{fig:time-intra}-(b) shows the time percentages of encoding sequences by HM with our complexity reduction approach vs. that by the original HM. Note that the results of Fig. \ref{fig:time-intra}-(b) are obtained by averaging over sequences with the same resolution (of 18 test sequences) encoded at four QP values $\{22,27,32,37\}$. As seen in this figure, ETH-CNN and ETH-LSTM achieve short running time compared with the encoding time of the original HM at the LDP configuration. Note that the time proportion of ETH-CNN and ETH-LSTM in inter-mode HEVC is far less than that of ETH-CNN in intra-mode HEVC, since inter-mode HEVC has a significantly longer encoding time than intra-mode HEVC. Besides, Fig. \ref{fig:time-intra}-(b) shows that the CU partition by our approach consumes relatively little encoding time, when applying our deep learning approach for complexity reduction in inter-mode HEVC.

As seen in Fig. \ref{fig:time-intra}-(b), ETH-CNN and ETH-LSTM achieve shorter running time compared with the encoding time of the original HM at the LDP configuration. Note that the time proportion of ETH-CNN and ETH-LSTM in inter-mode HEVC is far less than that of ETH-CNN in intra-mode HEVC, since inter-mode HEVC has a significantly longer encoding time than intra-mode HEVC. Besides, Fig. \ref{fig:time-intra}-(b) shows that the CU partition by our approach consumes relatively little encoding time, when applying our deep learning approach for complexity reduction in inter-mode HEVC.

%In addition, the number in blue font next to each bar in Fig.10 indicates the absolute running time of ETH-CNN and ETH-CNN + ETH-LSTM (in millisecond) for one frame. Here, the results of the absolute running time are obtained by averaging over all test images/sequences with the same resolution at four QP values $\{22,27,32,37\}$.
%Note that only one thread of CPU is used for the above running time, in accord with the HM platform that is also implemented in one thread.
%If enabling the multi-thread parallelism of CPU, the running time can be drastically reduced.

\subsection{Ablation Study}

\begin{table}
	\newcommand{\tabincell}[2]{\begin{tabular}{@{}#1@{}}#2\end{tabular}}
	\begin{center}
		\caption{\footnotesize{Results for ablation study}}
		\label{tab:ablation}
		\tiny
		\begin{tabular}{|c|c|c|c|c|c|c|c|}
			\hline \multirow{2}{*}{\hspace{-0.5em}Config.\hspace{-0.5em}} & \multirow{2}{*}{Settings} & \multirow{2}{*}{\tabincell{c}{\hspace{-0.5em}BD-BR\hspace{-0.5em} \\(\%)}} & \multirow{2}{*}{\tabincell{c}{\hspace{-0.5em}BD-PSNR\hspace{-0.5em} \\(dB)}} & \multicolumn{4}{|c|}{$\Delta{}T$ (\%)} \\
			\cline{5-8} & & & & QP=22 & QP=27 & QP=32 & QP=37 \\
			
			\hline \multirow{3}{*}{AI} & 1. Shallow CNN \cite{Liu16TIP} & 6.189 & -0.316 & -56.41 & -60.23 & -62.62 & -65.04 \\
			& 2. Deep CNN \cite{Li17ICME} & 2.249 & -0.105 & -55.44 & -58.23 & -60.12 & -63.79 \\
			& 3. ETH-CNN & \textbf{2.247} & \textbf{-0.104} & \textbf{-56.92} & \textbf{-60.38} & \textbf{-63.61} & \textbf{-66.47} \\
			
			\hline \multirow{7}{*}{LDP} & \multirow{2}{*}{\tabincell{c}{1. ETH-CNN fed with\\original CTUs}} & \multirow{2}{*}{8.911} & \multirow{2}{*}{-0.248} & \multirow{2}{*}{\textbf{-47.00}} & \multirow{2}{*}{\textbf{-55.10}} & \multirow{2}{*}{\textbf{-60.68}} & \multirow{2}{*}{\textbf{-65.53}} \\
			& & & & & & &\\
			& \multirow{2}{*}{\tabincell{c}{2. ETH-CNN fed with\\residual CTUs}} & \multirow{2}{*}{3.450} & \multirow{2}{*}{-0.106} & \multirow{2}{*}{-44.76} & \multirow{2}{*}{-52.87} & \multirow{2}{*}{-58.54} & \multirow{2}{*}{-63.79} \\
			& & & & & & & \\
			& \multirow{2}{*}{\tabincell{c}{3. ETH-CNN fed with\\residual CTUs + ETH-LSTM}} & \multirow{2}{*}{\textbf{1.495}} & \multirow{2}{*}{\textbf{-0.046}} & \multirow{2}{*}{-43.84} & \multirow{2}{*}{-52.13} & \multirow{2}{*}{-57.89} & \multirow{2}{*}{-62.94} \\
			& & & & & & &\\
			\hline
		\end{tabular}
	\end{center}
\end{table}

In this section, we conducted a series of ablation experiments to analyze the impact of major components in the proposed approach.
In our ablation experiments, we started from a simple CNN model and then added the components stepwise, finally reaching the proposed approach. Table \ref{tab:ablation} reports the results of ablation experiments. Note that the ablation results are averaged over all 18 test sequences. In the following, we discuss these ablation results in details.

\textbf{Deep CNN structure \textit{vs.} shallow CNN structure.}
In our approach, a deep structure of CNN is developed, which is significantly deeper than the shallow CNN structure in the state-of-the-art HEVC complexity reduction approach \cite{Liu16TIP}. Such a deep CNN structure has been presented in the conference version \cite{Li17ICME} of this paper, and it has similar structure and comparable amount of trainable parameters to our ETH-CNN model.
Therefore, we analyze the impact of the deep structure in our approach by comparing the results between \cite{Li17ICME} and \cite{Liu16TIP}.
As can be seen in Table \ref{tab:ablation}, the deep CNN structure improves the RD performance for intra-mode HEVC complexity reduction, which has $3.940\%$ BD-BR saving and $0.211$dB BD-PSNR increase over the shallow structure of \cite{Liu16TIP}.

\textbf{The ETH-CNN model \textit{vs.} the general classification-oriented CNN model.}
In the ETH-CNN model, the early-terminated hierarchical (ETH) decision is proposed to predict the CU partition, which replaces the conventional three-level classification of \cite{Li17ICME}.
In fact, \cite{Li17ICME} is the conference version of this paper, in which three deep networks of the CNN model are invoked to predict the three-level classification of $64\times64$, $32\times32$ and $16\times16$ CU partition, respectively.
Note that the CNN model of \cite{Li17ICME} has similar structure and comparable amount of trainable parameters to our ETH-CNN model.
In our ablation experiments, we therefore compare the results of the ETH-CNN model with those of the classification-oriented CNN model  \cite{Li17ICME}, to investigate the impact of the ETH decision.
We can see from Table \ref{tab:ablation} that the ETH-CNN model can save $1.48\% - 3.49\%$ encoding time over \cite{Li17ICME} for intra-mode HEVC.
This indicates the effectiveness of the ETH decision proposed in this paper, which replaces the general CNN model based classification in \cite{Li17ICME}.

\textbf{Residual CTUs \textit{vs.} original CTUs.}
For inter-mode HEVC complexity reduction, we propose to feed the ETH-CNN model with residual CTUs, rather than the original CTUs.
In our ablation experiments, we thus validate the effectiveness of residual CTUs in our approach.
To this end, we compare the performance of the ETH-CNN model fed with residual CTUs and original CTUs.
Note that original CTUs are replaced by residual CTUs in both training and test procedures.
Table \ref{tab:ablation} shows that the residual CTUs can improve the RD performance with $5.461\%$ BD-BR saving and $0.142$dB BD-PSNR increase.
The cost is $1.12\% - 2.24\%$ increase in the encoding time.

\textbf{ETH-CNN + LSTM \textit{vs.} only ETH-CNN.}
In our approach, ETH-LSTM is a major contribution for inter-mode HEVC complexity reduction. We thus show the performance improvement of embedding the ETH-LSTM model in our approach, i.e., the combination of the ETH-CNN and ETH-LSTM models. Note that the residual CTUs are as the input to the ETH-CNN model followed by the ETH-LSTM model. As shown in Table \ref{tab:ablation}, the ETH-LSTM model prominently improves the RD performance with less than one half BD-BR and BD-PSNR, and meanwhile has insignificant change of the encoding time. This indicates the effectiveness of the ETH-LSTM model in our approach. Besides, the ETH-LSTM model introduces little computational time in our approach, since Sections \ref{sec:complexity-cnn} and \ref{sec:complexity-lstm} have investigated that the ETH-LSTM model consumes much less floating-point operations than the ETH-CNN model.

\subsection{Performance Evaluation with Various Settings}

%In this section, we extended the experiments to evaluate the adaption of our approach with various settings.

\begin{figure}
	\centering
	\begin{minipage}{0.49\linewidth}
		\centering
		\includegraphics[width=1\linewidth]{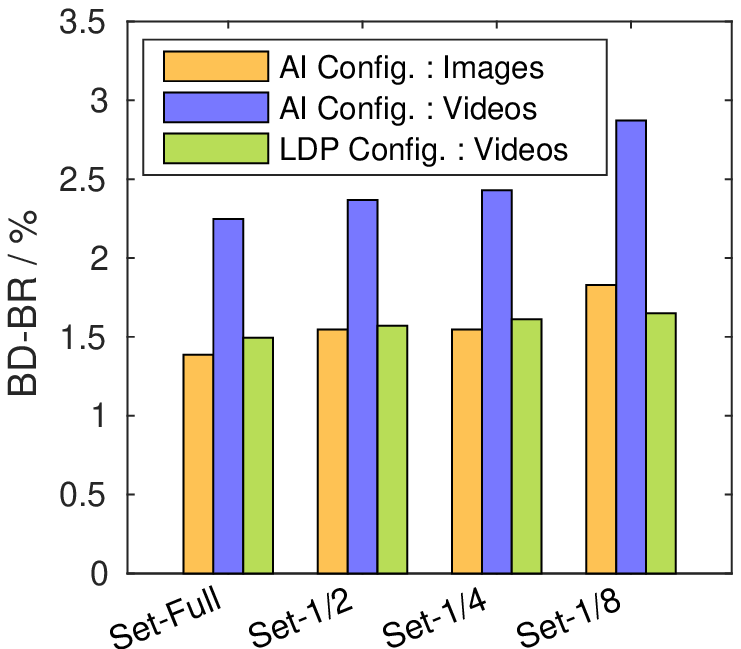}\\
		%\centerline{\footnotesize{(a) BD-BR}}\medskip
	\end{minipage}
	\hfill
	\begin{minipage}{0.49\linewidth}
		\centering
		\includegraphics[width=1\linewidth]{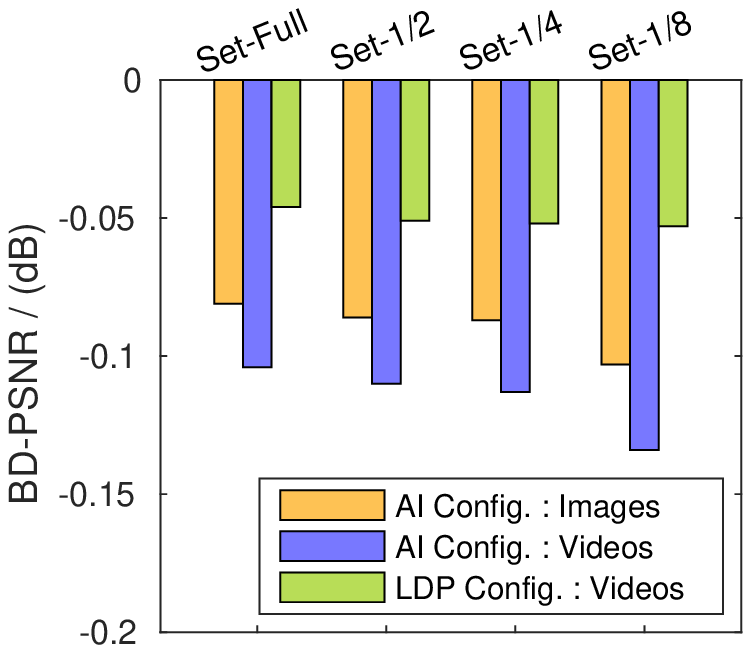}\\
		%\centerline{\footnotesize{(b) BD-PSNR}}\medskip
	\end{minipage}
	\caption{\footnotesize RD performance with different training sets. At the AI configuration, Set-$1/2$, Set-$1/4$ and Set-$1/8$ contain 850, 425 and 213 images, respectively. At the LDP configuration, those are 42, 21 and 11 video sequences.}\label{fig:diff-training}
\end{figure}

\textbf{Evaluation with different amount of training data.}
Since the proposed approach is data-driven, it is necessary to evaluate its performance when changing the amount of training data.
For intra- and inter-modes, the 1700 training images and 83 training sequences in our CPH-Intra and CPH-Inter databases are regarded as the full training sets, namely Set-Full.
For either intra- or inter-mode, three smaller training sets are generated via randomly selecting $1/2$, $1/4$ and $1/8$ of images or videos from the corresponding Set-Full, namely Set-$1/2$, Set-$1/4$ and Set-$1/8$, respectively.
Afterwards, the ETH-CNN and ETH-LSTM models are retrained on Set-$1/2$, Set-$1/4$ and Set-$1/8$, respectively.
Fig. \ref{fig:diff-training} shows the RD performance of our approach, when changing the amount of training data (i.e., Set-full, Set-$1/2$, Set-$1/4$ and Set-$1/8$).
Note that the results are averaged over all test images or all test sequences at QPs $22, 27, 32$ and $37$.
We can observe in Fig. \ref{fig:diff-training} that BD-BR decreases and BD-PSNR increases along with the increased amount of training data, for both AI (i.e., intra-mode) and LDP (i.e., inter-mode) configurations.
Therefore, the large-scale CPH databases are essential to achieve the desirable RD performance.

\begin{figure}
	\centering
	\begin{minipage}{0.32\linewidth}
		\centering
		\includegraphics[width=1.05\linewidth]{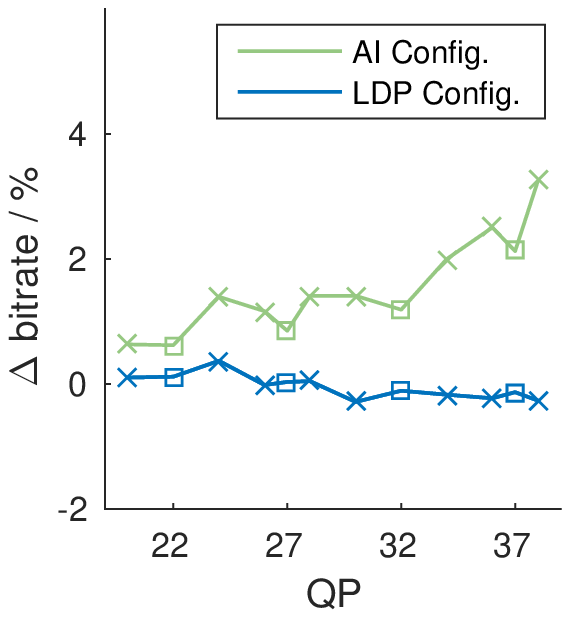}\\
		%\centerline{\footnotesize{(a) $\Delta$bitrate}}\medskip
	\end{minipage}
	\hfill
	\begin{minipage}{0.32\linewidth}
		\centering
		\includegraphics[width=1.05\linewidth]{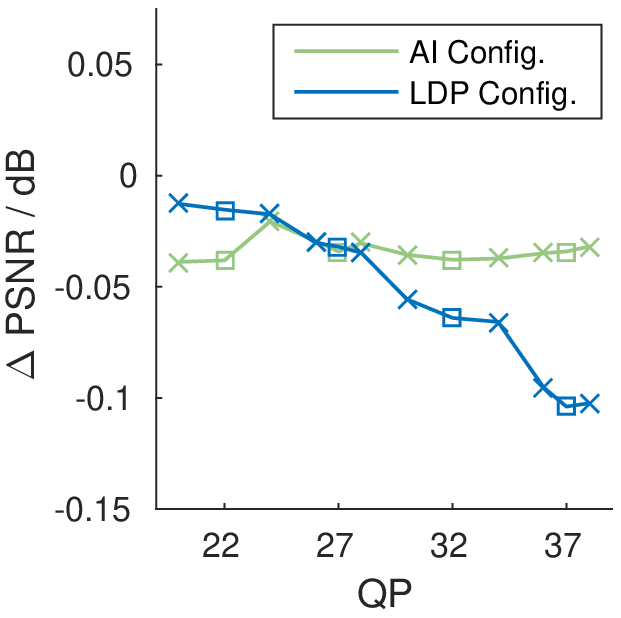}\\
		%\centerline{\footnotesize{(b) $\Delta$PSNR}}\medskip
	\end{minipage}
	\hfill
	\begin{minipage}{0.32\linewidth}
		\centering
		\includegraphics[width=1.05\linewidth]{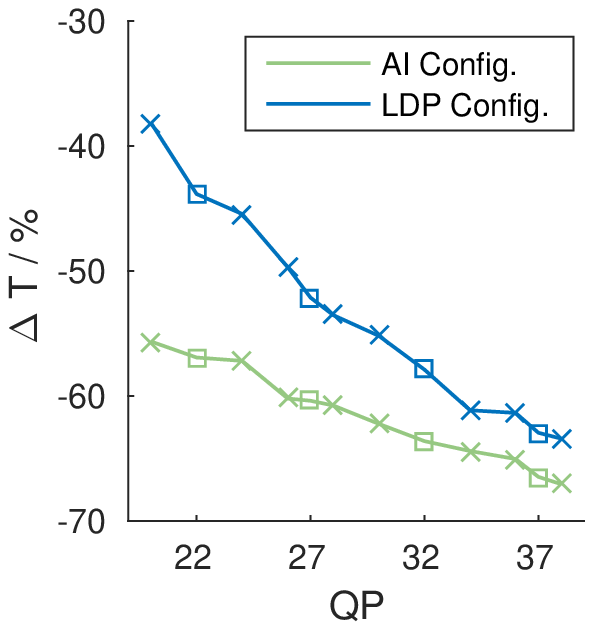}\\
		%\centerline{\footnotesize{(c) $\Delta{}T$}}\medskip
	\end{minipage}
	\caption{Curves for $\Delta$bitrate, $\Delta$PSNR and $\Delta{}T$ of our approach at different QPs. The square marks are the results at four anchored QPs, and the cross marks represent the results at other QPs.}
	\label{fig:diff-qp}
\end{figure}

\textbf{Generalization capability at different QPs.}
In addition to four QPs evaluated above (QP = 22, 27, 32, 37), we further test our approach for reducing complexity of intra- and inter-mode HEVC at other eight QPs, i.e., QP = $20, 24, 26, 28, 30, 34, 36, 38$.
To test the generalization capability of our approach, we directly use the ETH-CNN and ETH-LSTM models trained at four original QPs (i.e., QP = 22, 27, 32, 37) without re-training on the sequences compressed at other QPs.
Fig. \ref{fig:diff-qp} illustrates the bit-rate difference ($\Delta$bitrate), PSNR loss ($\Delta$PSNR) and time reduction ($\Delta{}T$) of our approach at different QPs, for complexity reduction of HEVC at both AI and LDP configurations. Note that the results are averaged over all test video sequences. In this figure, the square marks denote the test results at four original QPs, whereas the cross marks mean the test results at eight other QPs (the sequences compressed at these QPs are not used for training). As shown in Fig. \ref{fig:diff-qp}-left and -middle, $\Delta$bitrate and $\Delta$PSNR at each of other QPs (i.e., cross marks) are close to those of the adjacent original QPs (i.e., square marks), with only slight fluctuation. This indicates the generalization capability of our approach at different QPs. Besides, we can see from Fig. \ref{fig:diff-qp}-right that the higher QPs result in larger reduction rate of encoding time.

\begin{table}
	
	\newcommand{\tabincell}[2]{\begin{tabular}{@{}#1@{}}#2\end{tabular}}
	\begin{center}
		\caption{\footnotesize{Results for sequences of the JCT-VT test set (LDB \& RA)}}
		\label{tab:result-LDB-RA}
		\tiny
		\begin{tabular}{|c|c|c|c|c|c|c|c|}
			\hline \multirow{2}{*}{Config.} & \multirow{2}{*}{Approach} & \multirow{2}{*}{\tabincell{c}{\hspace{-0.5em}BD-BR\hspace{-0.5em} \\(\%)}} & \multirow{2}{*}{\tabincell{c}{\hspace{-0.5em}BD-PSNR\hspace{-0.5em} \\(dB)}} & \multicolumn{4}{|c|}{$\Delta{}T$ (\%)} \\
			\cline{5-8} & & & & QP=22 & QP=27 & QP=32 & QP=37 \\
			
			\hline \multirow{4}{*}{LDB} & \cite{Zhang15TIP} & 1.840 & -0.057 & -29.57 & -45.69 & -44.91 & -44.12 \\
			& \cite{Correa15TCSVT} & 4.312 & -0.144 & -32.07 & -38.68 & -48.03 & -57.11 \\
			& \cite{Mallikarachchi16TCSVT} & 3.458 & -0.109 & \textbf{-45.76} & -46.98 & -47.67 & -48.99 \\
			& Our & \textbf{1.722} & \textbf{-0.053} & -39.76 & \textbf{-47.43} & \textbf{-54.72} & \textbf{-59.84} \\
			
			\hline \multirow{4}{*}{RA} & \cite{Zhang15TIP} & 2.319 & -0.074 & -42.50 & -48.17 & -51.89 & -51.13 \\
			& \cite{Correa15TCSVT} & 6.246 & -0.210 & -40.96 & -49.17 & -55.41 & -61.10 \\
			& \cite{Mallikarachchi16TCSVT} & 6.101 & -0.204 & -40.17 & -41.98 & -43.20 & -43.02 \\
			& Our & \textbf{1.483} & \textbf{-0.048} & \textbf{-43.14} & \textbf{-51.64} & \textbf{-59.55} & \textbf{-64.07} \\
			\hline
		\end{tabular}
	\end{center}
\end{table}

\textbf{Evaluation at LDB and RA configurations.}
We further evaluate the performance of our approach for HEVC complexity reduction at the LDB and RA configurations. In our experiments, the ETH-CNN and ETH-LSTM models for both configurations were retrained from the training sequences at the corresponding configurations in our CPH-Inter database. For the LDB configuration, the experimental settings followed those for the LDP configuration. For the RA configuration, we slightly modified our approach to reduce HEVC complexity.
Specifically, the length of ETH-LSTM was 32 for both training and test at the RA configuration, in accord with the period between two adjacent I frames.
Here, the training samples were non-overlapping instead of the 10-frame overlapping used in the LDP configuration, ensuring all frames of the training samples locate in the same period within two adjacent random access points.
Besides, the order of frames fed into ETH-LSTM followed the order of encoding rather than that of displaying.
Table \ref{tab:result-LDB-RA} shows the performance of our and other state-of-the-art approaches for HEVC complexity reduction at the LDB and RA configurations.
Similar to the above evaluation, the results in this table are averaged over all 18 standard video sequences from the JCT-VC test set \cite{Correa12TCSVT}.
As shown in Table \ref{tab:result-LDB-RA}, our approach at the RA configuration is able to reduce $43.14\%\sim64.07\%$ encoding time while incurring $1.483\%$ of BD-BR increment and $0.048$dB of BD-PSNR loss, similar to those at the LDP configuration ($-43.84\%\sim-62.94\%$ of $\Delta{}T$, $1.495\%$ of BD-BR and $-0.046$dB of BD-PSNR).
More importantly, our approach outperforms other state-of-the-art approaches \cite{Zhang15TIP}, \cite{Correa15TCSVT} and \cite{Mallikarachchi16TCSVT} in terms of both encoding time and RD performance. At the LDB configuration, similar results can be found.
Therefore, the effectiveness of our approach at the LDB and RA configurations is verified.

\textbf{Evaluation on sequences with drastic scene change.}
In practice, drastic scene change may occur in videos.
Thus, we further evaluate the performance of our approach on the sequences that have scene change, under the LDP and RA configurations.
Among the 18 JCT-VC test sequences \cite{Correa12TCSVT}, only sequence \textit{Kimono} (1920$\times$1080) has drastic scene change. Hence, in addition to \textit{Kimono}, two sequences with drastic scene change, \textit{Tennis} (1920$\times$1080) and \textit{Mobisode} (832$\times$480), were tested in our experiments.
Table IV of the \textit{Supporting Document} shows the overall performance of our and other approaches, averaged over all frames of each sequence at QP $= 22, 27, 32$, and $37$.
We can see from this table that our approach outperforms other approaches for sequences with scene change.
Moreover, in the \textit{Supporting Document} we further compare the results of our and other approaches in the scene changed frames of the above three sequences, and the results show that our approach is robust to scene change.

\section{Conclusions}
\label{sec:conc}
In this paper, we have proposed a deep learning approach to reduce the encoding complexity of intra- and inter-mode HEVC, which learns to predict the optimal CU partition instead of using conventional brute-force RDO search. To deepen the networks of our deep learning approach, the CPH-Intra and CPH-Inter database was established, consisting of large-scale data of HEVC CU partition at intra- and inter modes, respectively. Then, two deep learning architectures, i.e., ETH-CNN and ETH-LSTM, were proposed to predict the CU partition for reducing the HEVC complexity at the intra- and inter-modes. The output of these architectures is HCPM, which hierarchically represents the CU partition in a CTU. Upon the representation of HCPM, an early termination mechanism was introduced in ETH-CNN and ETH-LSTM to save the computational time.
%Compared with the original HM, our approach averagely reduces encoding time by XXXX\% with negligible XXXX\% increased BD-BR for intra-mode HEVC, and by XXXX\% with negligible XXXX\% increased BD-BR for inter-mode HEVC.
The experimental results show that our deep learning approach performs much better than other state-of-the-art approaches in terms of both complexity reduction and RD performance.

There are three promising directions for future works. Our work, at the current stage, mainly focuses on predicting the CU partition to reduce the HEVC encoding complexity. Other components, such as PU and TU prediction, can also be replaced by deep learning models to further reduce the encoding complexity of HEVC. This is an interesting future work. In the deep learning area, various techniques have been proposed to accelerate the running speed of deep neural networks. Our deep learning approach may be sped up by applying these acceleration techniques, which is seen as another promising future work. Beyond the CPU implementation of the current work, our approach may be further implemented in the FPGA device in future, for practical applications.

\footnotesize
\bibliographystyle{IEEEtran}
\bibliography{bare_jrnl}

\end{document}